\documentclass{article}
\usepackage{arxiv}

\usepackage[utf8]{inputenc} 
\usepackage[T1]{fontenc}    
\usepackage{hyperref}       
\usepackage{url}            
\usepackage{booktabs}       
\usepackage{amsfonts}       
\usepackage{nicefrac}       
\usepackage{microtype}      
\usepackage{graphics, subfig, float}
\usepackage{epstopdf, epsfig} 
\usepackage{hyphenat}

\title{TGHop: An Explainable, Efficient and Lightweight Method
for Texture Generation}

\author{
  Xuejing Lei\\
  Media Communications Lab\\
  University of Southern California\\
  Los Angeles, CA, USA \\
  \texttt{xuejing@usc.edu} \\
  \And
  Ganning Zhao\\
  Media Communications Lab\\
  University of Southern California\\
  Los Angeles, CA, USA \\
  \texttt{ganningz@usc.edu} \\
  \And
  Kaitai Zhang\\
  Media Communications Lab\\
  University of Southern California\\
  Los Angeles, CA, USA \\
  \texttt{kaitaizh@usc.edu} \\
  \And
  C.-C. Jay Kuo\\
  Media Communications Lab\\
  University of Southern California\\
  Los Angeles, CA, USA \\
  \texttt{cckuo@usc.edu} \\
}

\begin{document}
\maketitle

\begin{abstract}
An explainable, efficient and lightweight method for texture generation,
called TGHop (an acronym of Texture Generation PixelHop), is proposed in
this work. Although synthesis of visually pleasant texture can be
achieved by deep neural networks, the associated models are large in
size, difficult to explain in theory, and computationally expensive in
training. In contrast, TGHop is small in its model size, mathematically
transparent, efficient in training and inference, and able to generate high
quality texture. Given an exemplary texture, TGHop first crops many
sample patches out of it to form a collection of sample patches called
the source.  Then, it analyzes pixel statistics of samples from the
source and obtains a sequence of fine-to-coarse subspaces for these
patches by using the PixelHop++ framework. To generate texture patches
with TGHop, we begin with the coarsest subspace, which is called the
core, and attempt to generate samples in each subspace by following the
distribution of real samples. Finally, texture patches are stitched to
form texture images of a large size. It is demonstrated by experimental
results that TGHop can generate texture images of superior quality with
a small model size and at a fast speed. 

\end{abstract}


\section{Introduction}

Automatic generation of visually pleasant texture that resembles exemplary texture has been studied for several decades since it is of theoretical interest in texture analysis and modeling. Research in texture generation benefits texture analysis and modeling research~\cite{tuceryan1993texture, chang1993texture, arivazhagan2003texture,zhu1998filters,zhu1997minimax,zhang2019texture,zhang2019data} by providing a perspective to understand the regularity and randomness of textures. Texture generation finds broad applications in computer graphics and computer vision, including visual special effects generation, digital image restoration, texture and image compression, etc. It is also closely related to image enhancement tasks including image de-noising and image super-resolution.

Early works of texture generation generates textures in pixel space. Based on exemplary input, texture can be generated pixel-by-pixel \cite{de1997multiresolution, efros1999texture, wei2000fast} or
patch-by-patch~\cite{efros2001image, liang2001real, cohen2003wang,kwatra2003graphcut, wu2004feature}, starting from a small unit and gradually growing to a larger image. These methods, however, suffer from slow generation time~\cite{efros1999texture, liang2001real} or limited diversity of generated textures~\cite{efros2001image, cohen2003wang, kwatra2005texture}. Later works transform texture images to a feature space with kernels and exploit the statistical correlation of features for texture generation. Commonly used kernels include the Gabor filters~\cite{heeger1995pyramid} and the steerable pyramid filter banks~\cite{portilla2000parametric}. This idea is still being actively studied with the the resurgence of neural networks. Various deep learning (DL) models, including Convolutional Neural Networks(CNNs) and Generative Adversarial Networks (GANs), yields visually pleasing results in texture generation. Compared to traditional methods, DL-based methods~\cite{gatys2015texture, liu2016texture,risser2017stable, li2017diversified, li2017universal,ustyuzhaninov2017does, shi2020fast} learn weights and biases through end-to-end optimization. Nevertheless, these models are usually large in model size, difficult to explain in theory, and computationally expensive in training. It is desired to develop a new generation method that is small in model
size, mathematically transparent, efficient in training and inference,
and able to offer high quality textures at the same time. Along this
line, we propose the TGHop (Texture Generation PixelHop) method in this
work.

TGHop consists of four steps. First, given an exemplary texture, TGHop
crops numerous sample patches out of it to form a collection of sample
patches called the source.  Second, it analyzes pixel statistics of
samples from the source and obtains a sequence of fine-to-coarse
subspaces for these patches by using the PixelHop++ framework~\cite{chen2020pixelhop++}. Third, to generate realistic
texture patches, it begins with generating samples in the coarsest subspace, which is called the core, by matching the distribution of real and generated samples, and attempts to generate spatial pixels given spectral coefficients from coarse to fine subspaces. Last, texture patches are stitched to form texture images of a larger size.  Extensive experiments are conducted to show that TGHop can generate texture images of superior quality with a small model size, at a fast speed, and in an explainable way. 

It is worthwhile to point out that this work is an extended version of
our previous work in \cite{lei2020nites}, where a method called NITES
was presented. Two works share the same core idea, but this work provides a more systematic study on texture synthesis task. In particular, a spatial Principal Component Analysis (PCA) transform is included in TGHop. This addition improves
the quality of generated textures and reduces the model size of TGHop as
compared with NITES.  Furthermore, more experimental results are given
to support our claim on efficiency (i.e., a faster computational speed)
and lightweight (i.e., a smaller model size). 

The rest of the paper is organized as follows. Related work is reviewed
in Sec. \ref{sec:review}. A high-level idea of successive subspace
analysis and generation is described in Sec. \ref{sec:SSL}. The TGHop
method is detailed in Sec.~\ref{sec:method}.  Experimental results are
shown in Sec.~\ref{sec:experiments}. Finally, concluding remarks and
future research directions are given in Sec.~\ref{sec:conclusion}. 

\section{Related Work}\label{sec:review}

\subsection{Early Work on Texture Generation}

Texture generation (or synthesis) has been a long-standing problem of great interest. The methods for it can be categorized into two types.  The first
type generates one pixel or one patch at a time and grows synthesized
texture from small to large regions. Pixel-based method synthesizes a
center pixel conditioned on its neighboring pixels. Efros and Leung~\cite{efros1999texture} proposed to synthesize a pixel by randomly choosing from the pixels that have similar neighborhood as the query pixel. Patch-based
methods~\cite{efros2001image, liang2001real, cohen2003wang,kwatra2003graphcut, wu2004feature} usually achieves higher quality than pixel-based methods~\cite{de1997multiresolution,efros1999texture,wei2000fast}. They suffer from two problems. First, searching the whole space to find a matched patch is slow~\cite{efros1999texture, liang2001real}. Second, the methods~\cite{efros2001image, cohen2003wang, kwatra2005texture} that stitching small patches to form a larger image sustain limited diversity of generated patches, though they are capable of producing high quality textures at a fast speed. A certain pattern may repeat several times in these generated textures without sufficient variations due to lack of understanding the perceptual properties of texture images. The second type addresses this problem by analyzing textures in feature spaces rather than pixel space. A texture image is first transformed to a feature space with kernels. Then, statistics in the feature space, such as histograms
\cite{heeger1995pyramid} and handcrafted summary
\cite{portilla2000parametric}, is analyzed and exploited for texture
generation. For the transform, some pre-defined filters such as Gabor
filters~\cite{heeger1995pyramid} or steerable pyramid filter
banks~\cite{portilla2000parametric} were adopted in early days. The design of these filters, however, heavily relies on human expertise and lack adaptivity. With the recent advances of deep neural networks, filters from a pre-trained networks such as VGG provide a powerful transformation for analyzing texture images and their statistics~\cite{gatys2015texture, li2017universal}.

\subsection{Deep-Learning-based (DL-based) Texture Generation}
DL-based methods often employ a texture loss function that computes the statistics of the features. Fixing the weights of a pre-trained network, the method in \cite{gatys2015texture} applies the Gram matrix as the statistical measurement and iteratively optimizes an initial white-noise input image through
back-propagation. The method in~\cite{li2017universal} computes feature covariances of white-noise image and texture image, and matches them through whitening and coloring. Both of these two methods utilized a VGG-19 network pre-trained on Imagenet dataset to extract features. The method in~\cite{ustyuzhaninov2017does} abandons the deep VGG network but adopt only one convolutional layer with random filter weights. Although these methods can generate visually pleasant textures, the iterative optimization process (i.e. backpropagation) is computationally expensive. There is a lot of follow-ups to \cite{gatys2015texture} such as incorporating other optimization terms~\cite{liu2016texture, risser2017stable} and improving inference speed~\cite{li2017diversified, shi2020fast}. However, there is a price to pay. The former aggravates the computational burden while the latter increases the training time. Another problem of these methods lies in the difficulty of explaining the usage of a pre-trained network. The methods in ~\cite{gatys2015texture, li2017universal} develop upon a VGG-19 network pre-trained on the Imagenet dataset. The Imagenet dataset is designed for understanding the semantic meaning of a large number of natural images. Textures, however, mainly contains low-level image characteristics. Although shallow layers (such as conv\_1) of VGG are known to capture low-level characteristics of images, generating texture only with shallow layers does not give a good results in~\cite{gatys2015texture}. It is hard to justify whether the VGG feature contains redundancy for textures or ignores some texture-specific information. Lack of explainability also raises the challenge of inspecting the methods when unexpected generation results occurred. Thus, these drawbacks motivate us to design a method that is efficient, lightweight and dedicated to texture.

\subsection{Successive Subspace Learning (SSL)}

To reduce the computational burden in training and inference of
DL-based methods, we adopt spatial-spectral representations for texture
images based on the successive subspace learning (SSL) framework
~\cite{kuo2016understanding, kuo2017cnn, kuo2019interpretable}. To implement SSL, PixelHop~\cite{chen2020pixelhop}
and PixelHop++~\cite{chen2020pixelhop++} architectures have been
developed.  PixelHop consists of multi-stage Saab transforms in cascade.
PixelHop++ is an improved version of PixelHop by replacing the Saab
transform with the channel-wise (c/w) Saab transform, exploiting weak correlations among spectral channels. Both the Saab
transform and the c/w Saab transform are data-driven transforms, which are variants of the PCA transform.  PixelHop++ offers powerful
hierarchical representations and plays a key role of dimension reduction in TGHop. SSL-based solutions have been proposed to tackle quite a few problems,
including~\cite{chen2021defakehop, zhang2021anomalyhop, kadam2021r,liu2021voxelhop, zhang2020pointhop++, zhang2020pointhop,zhang2020unsupervised,kadam2020unsupervised,manimaran2020visualization, tseng2020interpretable,rouhsedaghat2020facehop, rouhsedaghat2021successive}. In this work, we present an SSL-based texture image generation method. The main idea of the method is demonstrated in the next section. 

\section{Successive Subspace Analysis and Generation}\label{sec:SSL}

\begin{figure*}[t!]
\centering
\includegraphics[width=\linewidth]{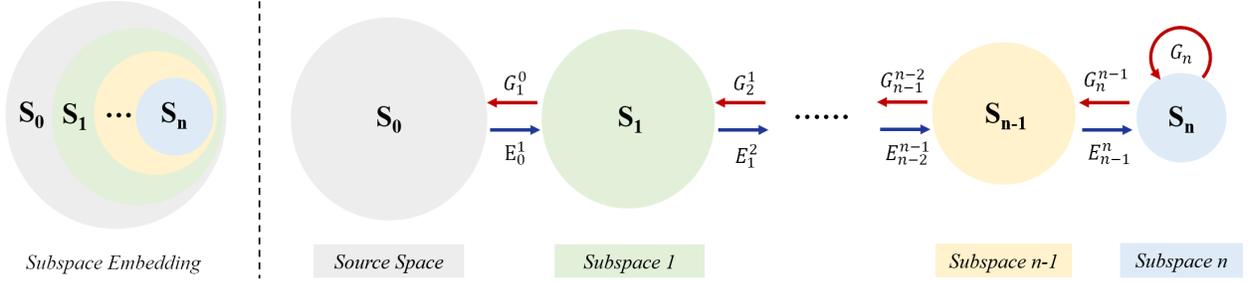}
\caption{Illustration of successive subspace analysis and generation,
where a sequence of subspace $S_1,\dots, S_n$ is constructed from source
space, $S_0$, through a successive process indicated by blue arrows
while red arrows indicate the successive subspace generation
process.}\label{fig:subspace}
\end{figure*}

In this section, we explain the main idea behind the TGHop method,
successive subspace analysis and generation, as illustrated in Fig.
\ref{fig:subspace}.  Consider an input signal space denoted by
$\tilde{S}_0$, and a sequence of subspaces denoted by $\tilde{S}_1,
\cdots, \tilde{S}_n$. Their dimensions are denoted by $\tilde{D}_0$,
$\tilde{D}_1$, $\cdots$, $\tilde{D}_n$.  They are related with each
other by the constraint that any element in $\tilde{S}_{i+1}$ is formed by
an affine combination of elements in $\tilde{S}_{i}$, where $i=0,
\cdots, n-1$. 

An affine transform can be converted to a linear transform by augmenting
vector $\tilde{{\bf a}}$ in $\tilde{S}_{i}$ via ${\bf a}=(\tilde{{\bf
a}}^T, 1)^T$. We use $S_i$ to denote the augmented space of
$\tilde{S}_{i}$ and $D_i=\tilde{D}_i+1$. Then, we have the following relationship
\begin{equation}\label{eq:embedding}
S_n \subset S_{n-1} \subset \cdots \subset S_1 \subset S_0,
\end{equation}
and
\begin{equation}\label{eq:dim}
D_n < D_{n-1} < \cdots <  D_1 < D_0.
\end{equation}

We use texture analysis and generation as an example to explain this
pipeline. To generate homogeneous texture, we collect a number of texture
patches cropped out of exemplary texture as the input set.  Suppose that
each texture patch has three RGB color channels, and a spatial
resolution $P\times P$. The input set then has a dimension of $3 P^2$
and its augmented space $S_0$ has a dimension of $D_0=3 P^2 + 1$.  If
$P=32$, we have $D_0=3073$ which is too high to find an effective
generation model directly. 

To address this challenge, we build a sequence of subspaces
$S_0$, $S_1$, $\cdots$, $S_n$ with decreasing dimensions. We call $S_0$
and $S_n$ the "source" space and the "core" subspace, respectively.  We
need to find an effective subspace $S_{i+1}$ from $S_i$, and
such an analysis model is denoted by $E_i^{i+1}$. Proper subspace
analysis is important since it determines how to decompose an input
space into the direct sum of two subspaces in the forward analysis
path. Although we choose one of the two for further processing and discard the other one, we need to record the relationship of the two decomposed subspaces so that they are well-separated in the reverse generation path. This forward process is called fine-to-coarse analysis. 

In the reverse path, we begin with the generation of samples in $S_n$ by
studying its own statistics. This is accomplished by generation model
$G_n$. The process is called core sample generation. Then, conditioned on a generated sample in $S_{i+1}$, we generate a new sample in $S_{i}$ through a generation model denoted by $G_{i+1}^i$. This process is called coarse-to-fine generation. In Fig. \ref{fig:subspace}, we use blue and red arrows to indicate analysis and generation, respectively. This idea can be implemented as a non-parametric method since we can choose subspaces $S_1$, $\cdots$, $S_n$, flexibly in a feedforward manner. One specific design is elaborated in the next section. 

\section{TGHop Method}\label{sec:method}

The TGHop method is proposed in this section. An overview of the TGHop
method is given in Sec.~\ref{subsec:system}. Next, the forward
fine-to-coarse analysis based on the two-stage c/w Saab transforms is
discussed in Sec.~\ref{subsec:sse}. Afterwards, sample generation in the
core is elaborated in Sec.~\ref{subsec:csg}.  Finally, the reverse
coarse-to-fine pipeline is detailed in Sec.~\ref{subsec:ssg}. 

\subsection{System Overview}\label{subsec:system}

\begin{figure*}[ht]
\centering
\includegraphics[width=\linewidth]{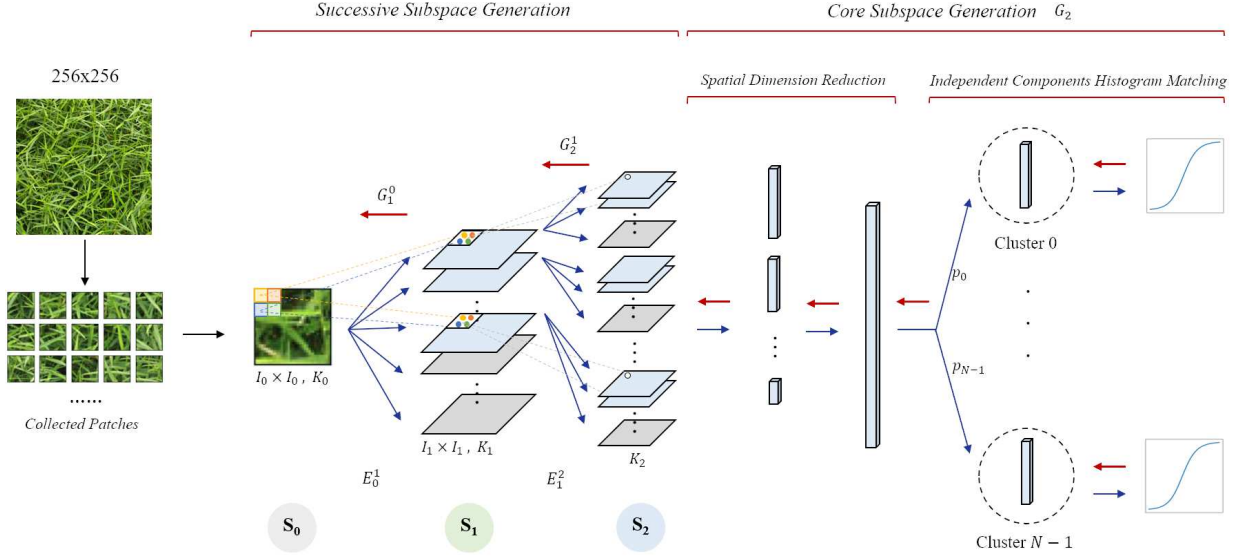}
\caption{An overview of the proposed TGHop method. A number of patches
are collected from the exemplary texture image, forming source space
$S_0$. Subspace $S_1$ and $S_2$ are constructed through analysis model
$E_0^1$ and $E_1^2$. Input filter window sizes to Hop-1 and Hop-2 are
denoted as $I_0$ and $I_1$. Selected channel numbers of Hop-1 and Hop-2
are denoted as $K_1$ and $K_2$. A block of size $I_i \times I_i$ of
$K_i$ channels in space/subspace $S_i$ is converted to the same spatial
location of $K_{i+1}$ channels in subspace $S_{i+1}$. Red arrows
indicate the generation process beginning from core sample generation
followed by coarse-to-fine generation. The model for core sample
generation is denoted as $G_2$ and the models for coarse-to-fine
generation are denoted as $G_2^1$ and $G_1^0$. }\label{fig:framework}
\end{figure*}

An overview of the TGHop method is given in Fig.~\ref{fig:framework}.
The exemplary color texture image has a spatial resolution of $256
\times 256$ and three RGB channels. We would like to generate multiple
texture images that are visually similar to the exemplary one.  By
randomly cropping patches of size $32 \times 32$ out of the source
image, we obtain a collection of texture patches serving as the input to
TGHop. The dimension of these patches is $32 \times 32 \times 3=3072$.
Their augmented vectors form source space $S_0$. The TGHop system is
designed to generate texture patches of the same size that are visually
similar to samples in $S_0$.  This is feasible if we can capture both
global and local patterns of these samples. There are two paths in
Fig.~\ref{fig:framework}. The blue arrows go from left to right,
denoting the fine-to-coarse analysis process. The red arrows go
from right to left, denoting the coarse-to-fine generation process.
We can generate as many texture patches as desired using this procedure.
In order to generate a texture image of a larger size, we perform image
quilting~\cite{efros2001image} based on synthesized patches. 

\subsection{Fine-to-Coarse Analysis}\label{subsec:sse}

The global structure of an image (or an image patch) can be well
characterized by spectral analysis, yet it is limited in capturing local
detail such as boundaries between regions. Joint spatial-spectral
representations offer an ideal solution to the description of both
global shape and local detail information.  Analysis model $E_0^1$ finds
a proper subspace, $S_1$, in $S_0$ while analysis model $E_1^2$ finds a
proper subspace, $S_2$, in $S_1$. As shown in Fig.~\ref{fig:framework},
TGHop applies two-stage transforms.  They correspond to $E_0^1$ and
$E_1^2$, respectively.  Specifically, we can apply the c/w Saab
transform in each stage to conduct the analysis.  In the following, we provide a brief review on the Saab transform~\cite{kuo2019interpretable} and the c/w Saab
transform~\cite{chen2020pixelhop++}. 

We partition each input patch into non-overlapping blocks, each of which
has a spatial resolution of $I_0 \times I_0$ with $K_0$ channels.  We
flatten 3D tensors into 1D vectors, and decompose each vector into the
sum of one Direct Current (DC) and multiple Alternating Current (AC) 
spectral components. The DC filter is a
all-ones filter weighted by a constant. AC filters are obtained by
applying the principal component analysis (PCA) to DC-removed residual
tensor.  By setting $I_0=2$ and $K_0=3$, we have a tensor block of
dimension $2 \times 2 \times 3=12$.  Filter responses of PCA can be
positive or negative. There is a sign confusion problem~\cite{kuo2016understanding,kuo2017cnn} if both of them are allowed to
enter the transform in the next stage. To avoid sign confusion, a
constant bias term is added to all filter responses to ensure that all
responses become positive, leading to the name of the "subspace
approximation with adjusted bias (Saab)" transform. The Saab transform
is a data-driven transform, which is significantly different from
traditional transforms (e.g. Fourier and wavelet transforms) which are
data independent.  We partition AC channels into two low- and high-frequency 
bands. The energy of high-frequency channels (shaded by gray color
in Fig.~\ref{fig:framework}) is low and they are discarded for dimension
reduction without affecting the performance much. The energy of
low-frequency channels (shaded by blue color in
Fig.~\ref{fig:framework}) is higher.  For a tensor of dimension 12, we
have one DC and 11 AC components. Typically, we select $K_1=6$ to 10
leading AC components and discard the rest. Thus, after $E_0^1$, one 12-D
tensor becomes a $K_1$-D vector, which is illustrated by dots in subspace
$S_1$.  The $K_1$-D response vectors are fed into the next stage for
another transform. 

The channel-wise (c/w) Saab transform \cite{chen2020pixelhop++} exploits
the weak correlation property between channels so that the Saab
transform can be applied to each channel separately (see the middle part
of Fig.~\ref{fig:framework}). The c/w Saab transform offers an improved 
version of the standard Saab transform with a smaller model size. 

One typical setting used in our experiments is shown below.
\begin{itemize}
    \item Dimension of the input patch ($\tilde{D}_0$): $32\times32\times3=3072$;
    \item Dimension of subspace $\tilde{S}_1$ ($\tilde{D}_1$):
    $16\times16\times10=2560$ (by keeping 10 channels in Hop-1);
    \item Dimension of subspace $\tilde{S}_2$ ($\tilde{D}_2$):
    $8\times8\times27=1728$ (by keeping 27 channels in Hop-2).
\end{itemize}
Note that the ratio between $\tilde{D}_1$ and $\tilde{D}_0$ is 83.3\%
while that between $\tilde{D}_2$ and $\tilde{D}_1$ is 67.5\%. We are
able to reduce the dimension of the source space to that of the core
subspace by a factor of 56.3\%. In the reverse path indicated by red
arrows, we need to develop a multi-stage generation process. It should
also be emphasized that users can flexibly choose channel numbers in Hop-1 and
Hop-2. Thus, TGHop is a non-parametric method. 

The first-stage Saab transform provides the spectral information on the
nearest neighborhood, which is the first hop of the center pixel. By
generalizing from one to multiple hops, we can capture the information
in the short-, mid- and long-range neighborhoods. This is analogous to
increasingly larger receptive fields in deeper layers of CNNs. However,
filter weights in CNNs are learned from end-to-end optimization via
backpropagation while weights of the Saab filters in different hops are
determined by a sequence of PCAs in a feedforward unsupervised manner. 

\subsection{Core Sample Generation}\label{subsec:csg}

\begin{figure}[tb]
    \centering
    \subfloat[\centering without SDR] {{\includegraphics[width=0.2\columnwidth]{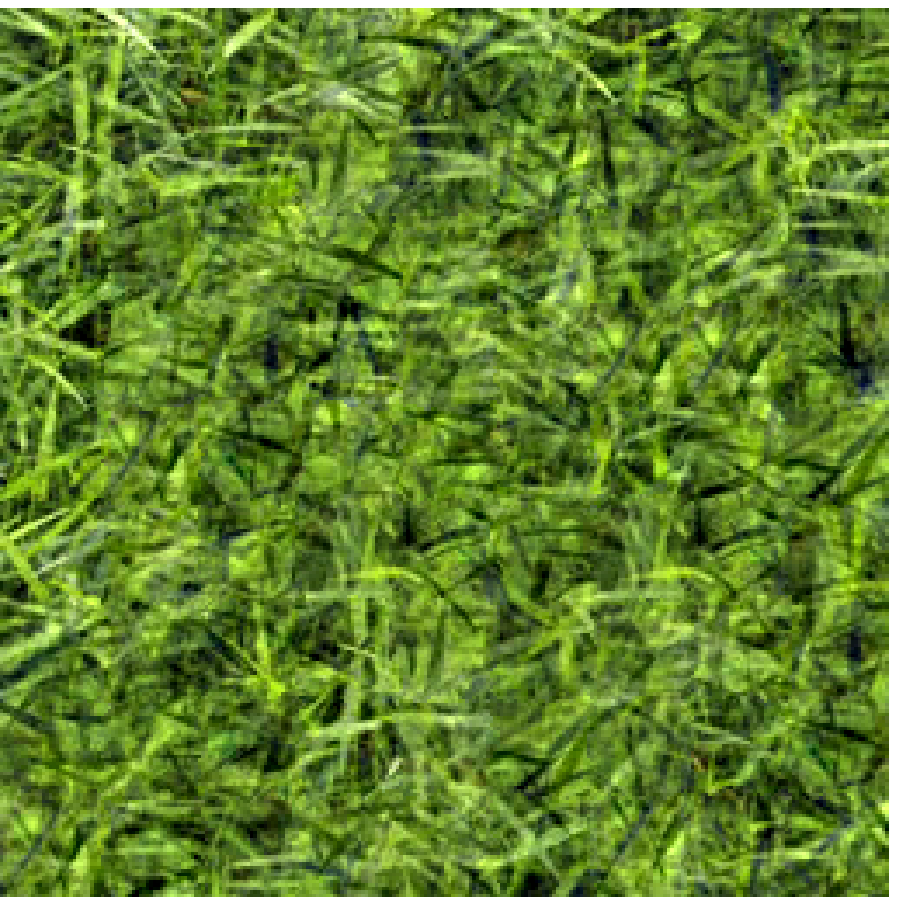} }}%
    \qquad
    \subfloat[\centering with SDR] {{\includegraphics[width=0.2\columnwidth]{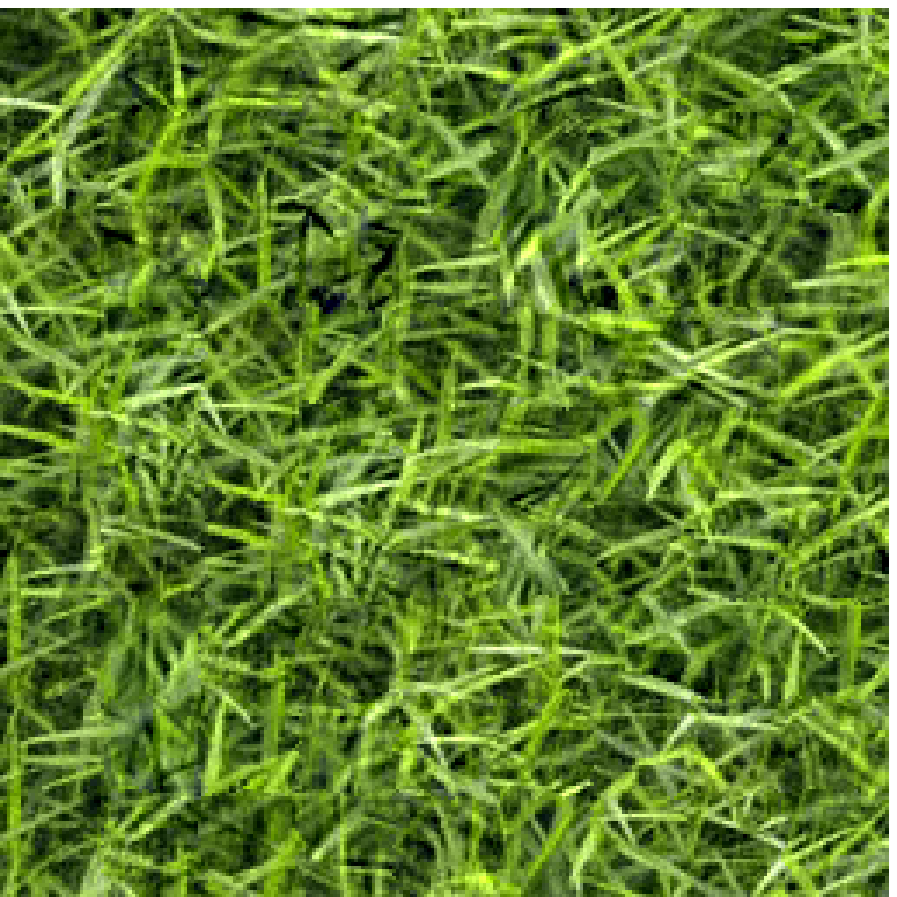} }}%
    \caption{Generated \emph{grass} texture image with and without spatial dimension reduction (SDR).}\label{fig:grass}%
\end{figure}

In the generation path, we begin with sample generation in core $S_n$
which is denoted by $G_n$. In the current design, $n=2$. We first
characterize the sample statistics in the core, $S_2$.  After two-stage
c/w Saab transforms, the sample dimension in $S_2$ is less than 2000.
Each sample contains $K_2$ channels of spatial dimension $8\times8$.
Since there exist correlations between spatial responses in each
channel, PCA is adopted for further Spatial Dimension Reduction (SDR).
We discard PCA components whose variances are lower than threshold
$\gamma$.  The same threshold applies to all channels.  SDR can help
reduce the model size and improve the quality of generated textures.
For example, we compare a generated \emph{grass} texture with and without
SDR in Fig.~\ref{fig:grass}. The quality with SDR significantly
improves. 

After SDR, we flatten the PCA responses of each channel and concatenate
them into a 1D vector denoted by $\mathbf{z}$. It is a sample in $S_2$.
To simplify the distribution characterization of a high-dimensional
random vector, we group training samples into clusters and transform
random vectors in each cluster into a set of independent random
variables.  We adopt the K-Means clustering algorithm to cluster
training samples into $N$ clusters, which are denoted by $\{C_i\}$,
$i=0,\cdots,N-1$.  Rather than modeling probability $P(\mathbf{z})$
directly, we model condition probability $P(\mathbf{z}\mid \mathbf{z}\in
C_i)$ with a fixed cluster index. The probability, $P(\mathbf{z})$, can
be written as
\begin{equation}\label{eq:prob}
P(\mathbf{z}) = \sum\limits_{i=0}^{N-1} P(\mathbf{z}\mid \mathbf{z}\in
C_i) \cdot P(\mathbf{z}\in C_i),
\end{equation}
where $P(\mathbf{z}\in C_i)$ is the percentage of data points in cluster
$C_i$. It is abbreviated as $p_i$, $i=0,\dots,N-1$ (see the right part of
Fig.~\ref{fig:framework}). 

Typically, a set of independent Gaussian random variables is used for
image generation. To do the same, we convert a collection of correlated
random vectors into a set of independent Gaussian random variables. To
achieve this objective, we transform random vector $\mathbf{z}$ in
cluster $C_i$ into a set of independent random variables through
independent component analysis (ICA), where non-Gaussianity serves as an
indicator of statistical independence. ICA finds applications in noise
reduction~\cite{hyvarinen1999sparse}, face
recognition~\cite{bartlett2002face}, and image
infusion~\cite{mitianoudis2007pixel}.  Our implementation is detailed
below. 
\begin{enumerate}
    \item Apply PCA to $\mathbf{z}$ in cluster $C_i$ for dimension reduction 
    and data whitening. 
    \item Apply FastICA~\cite{hyvarinen2000independent}, which is conceptually simple, 
    computationally efficient and robust to outliers, to the PCA output.
    \item Compute the cumulative density function (CDF) of each ICA
    component of random vector $\mathbf{z}$ in each cluster based on its
    histogram of training samples. 
    \item Match the CDF in Step 3 with the CDF of a Gaussian random variable
    (see the right part of Fig.  \ref{fig:framework}), where the inverse CDF
    is obtained by resampling between bins with linear interpolation.  To
    reduce the model size, we quantize N-dimensional CDFs, which have $N$
    bins, with vector quantization (VQ) and store the codebook of quantized CDFs. 
\end{enumerate}

\begin{figure}[htb]
\begin{center}
\includegraphics[width=1.45\linewidth]{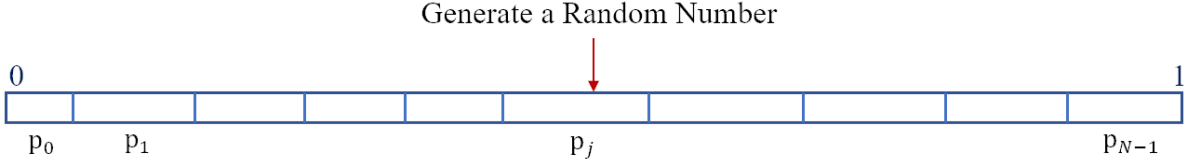}
\end{center}
\caption{Illustration of the interval representation, where the length of a
segment in the unit interval represents the probability of a cluster,
$p_i$. A random number is generated in the unit interval to indicate the
cluster index. }\label{fig:interval}
\end{figure}

We encode $p_i$ in Eq. (\ref{eq:prob}) using the length of a segment in
$[0,1]$. All segments are concatenated in order to build the unit
interval. The segment index is the cluster index. These segments are
called the interval representation as shown in Fig.~\ref{fig:interval}.
To draw a sample from subspace $S_2$, we use the uniform random number
generator to select a random number from interval $[0,1]$. This random
number indicates the cluster index on the interval representation. 

To generate a new sample in $S_2$, we perform the following steps:
\begin{enumerate}
    \item Select a random number from the uniform random number generator to determine the cluster index.
    \item Draw a set of samples independently from the Gaussian distribution.
    \item Match histograms of the generated Gaussian samples with the inverse CDFs in the chosen cluster.
    \item Repeat Steps 1-3 if the generated sample of Step 3 has no value larger than a pre-set threshold.
    \item Perform the inverse transform of ICA and the inverse transform of PCA.
    \item Reshape the 1D vector into a 3D tensor and this tensor is the generated sample in $S_2$.
\end{enumerate}

The above procedure is named Independent Components Histogram Matching
(ICHM). To conclude, there are two main modules in core sample
generation: spatial dimension reduction and independent components
histogram matching as shown in Fig.~\ref{fig:framework}. 

\subsection{Coarse-to-Fine Generation}\label{subsec:ssg}

\begin{figure}[tb]
\begin{center}
\includegraphics[width=0.7\linewidth]{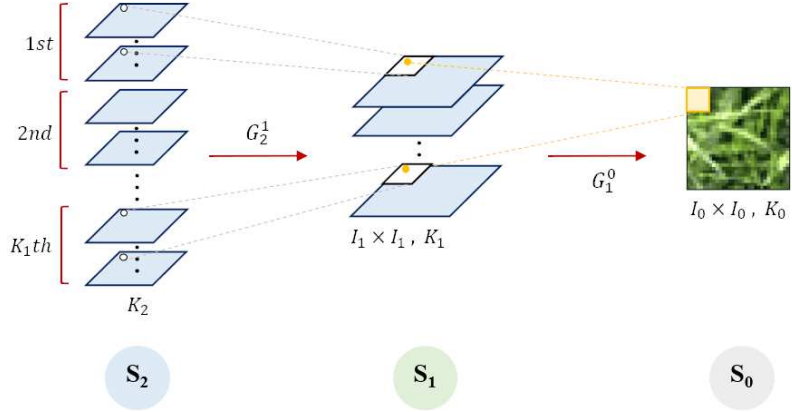}
\end{center}
\caption{Illustration of the generation process.}\label{fig:ssg}
\end{figure}

In this section, we examine generation model $G_{i+1}^i$, whose role is
to generate a sample in $S_i$ given a sample in $S_{i+1}$.  Analysis
model, $E_i^{i+1}$, transforms $S_i$ to $S_{i+1}$ through the c/w Saab
transform in the forward path.  In the reverse path, we perform the
inverse c/w Saab transform on generated samples in $S_{i+1}$ to $S_i$.
We take generation model $G_2^1$ as an example to explain the generation
process from $S_2$ and to $S_1$.  A generated sample in $S_2$ can be
partitioned into $K_1$ groups as shown in the left part of
Fig.~\ref{fig:ssg}. Each group of channels is composed of one DC channel
and several low-frequency AC channels. The $k$th group of channels in
$S_2$, whose number is denoted by $K_2^{(k)}$, is derived from the $k$th
channel in $S_1$. We apply the inverse c/w Saab transform to each group
individually. The inverse c/w Saab transform converts the tensor at the
same spatial location across $K_2^{(k)}$ channels (represented by white
dots in Fig.~\ref{fig:ssg}) in $S_2$ into a block of size $I_i \times
I_i$ (represented by the white parallelogram in Fig.~\ref{fig:ssg}) in
$S_1$, using the DC and AC components obtained in the fine-to-coarse analysis.  After the inverse c/w Saab transform, the
Saab coefficients in $S_1$ form a generated sample in $S_1$. The same
procedure is repeated between $S_1$ and $S_0$. 

\begin{figure}[htb]
    \centering
    \subfloat {{\includegraphics[width=0.21\linewidth]{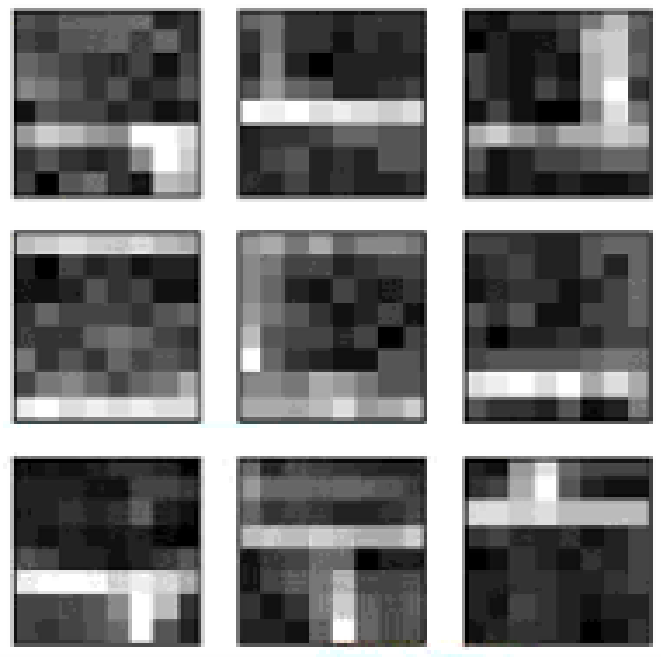} }}%
    \quad
    \subfloat {{\includegraphics[width=0.21\linewidth]{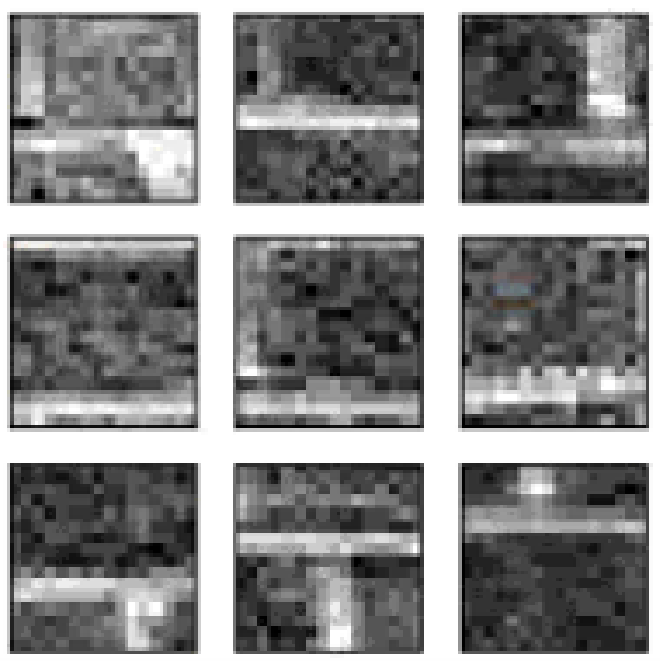} }}%
    \quad
    \subfloat {{\includegraphics[width=0.21\linewidth]{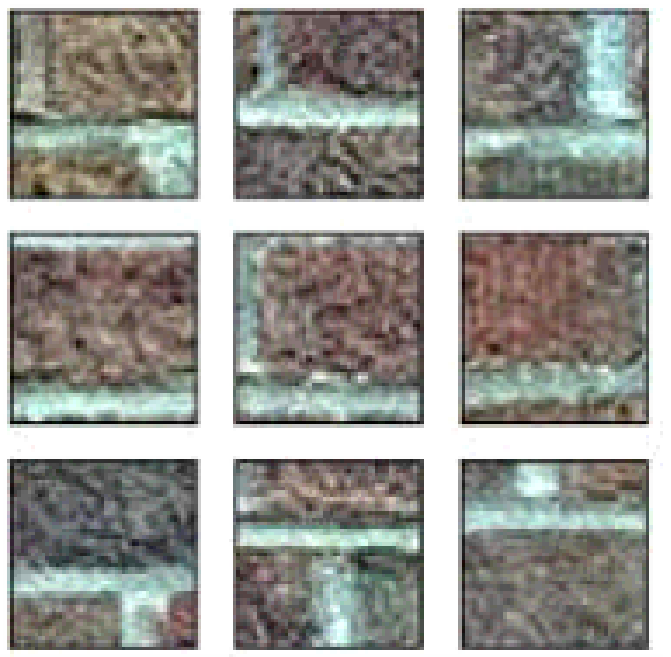} }}%

    \subfloat {{\includegraphics[width=0.21\linewidth]{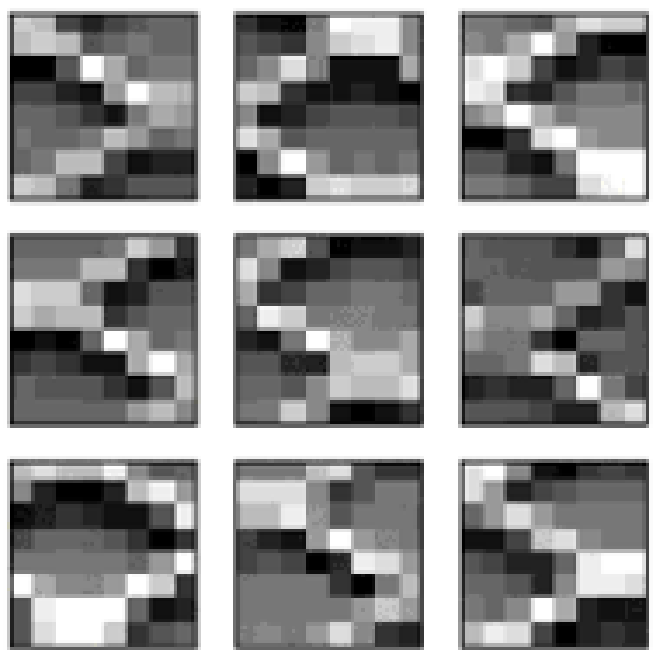} }}%
    \quad
    \subfloat {{\includegraphics[width=0.21\linewidth]{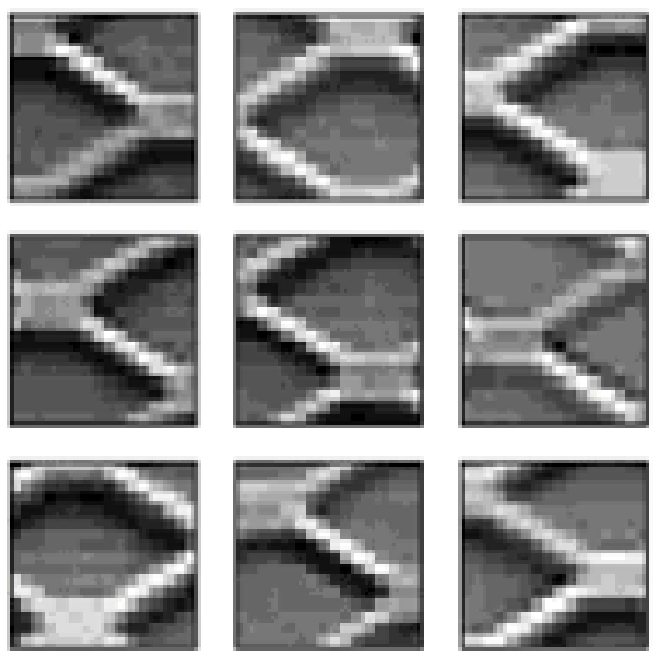} }}%
    \quad
    \subfloat {{\includegraphics[width=0.21\linewidth]{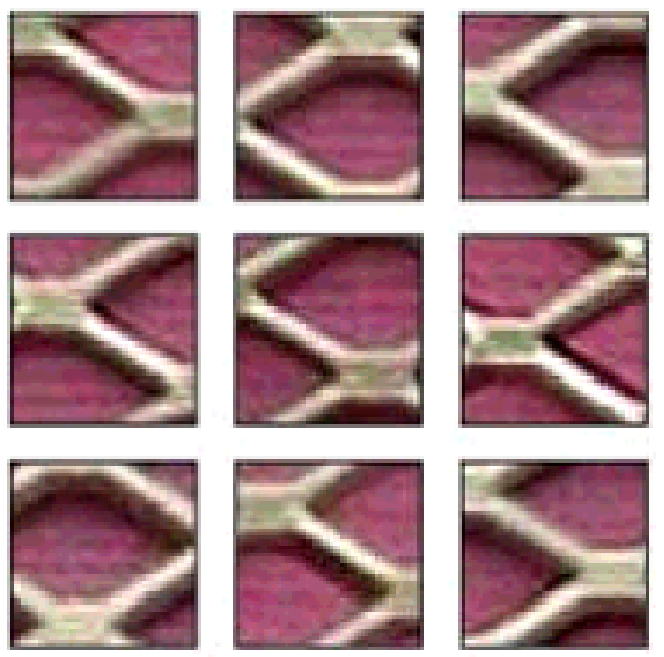} }}%
    
    \subfloat {{\includegraphics[width=0.21\linewidth]{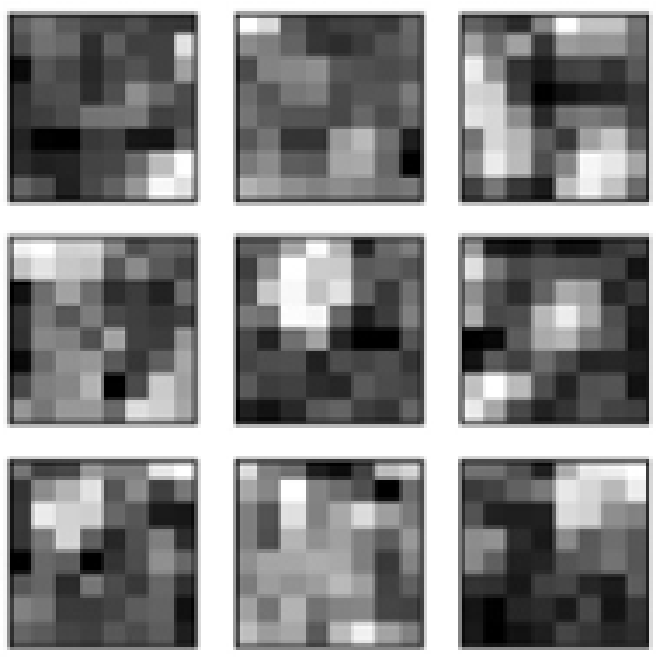} }}%
    \quad
    \subfloat {{\includegraphics[width=0.21\linewidth]{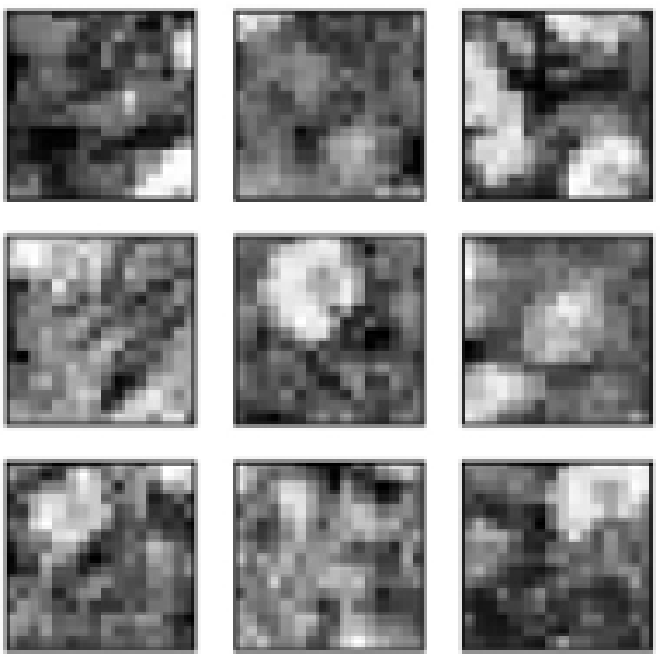} }}%
    \quad
    \subfloat {{\includegraphics[width=0.21\linewidth]{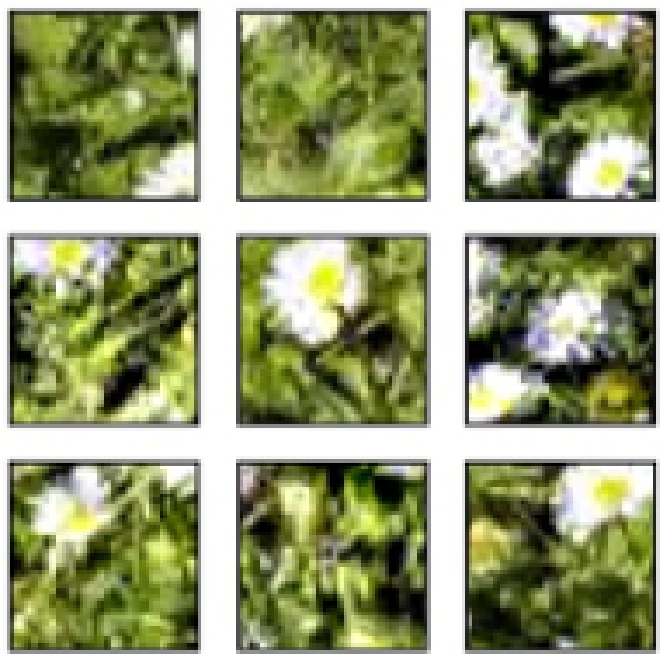} }}%
    \caption{Examples of generated DC maps in core $S_2$ (first column), generated
    samples in subspace $S_1$ (second co-lumn), and the ultimate generated textures 
    in source $S_0$ (third column).}%
    \label{fig:between}%
\end{figure}

Examples of several generated textures in core $S_2$, intermediate
subspace $S_1$ and source $S_0$ are shown in Fig.~\ref{fig:between}.
The DC channels generated in the core offer gray-scale low-resolution
patterns of a generated sample. More local details are added gradually
from $S_2$ to $S_1$ and from $S_1$ to $S_0$.

\section{Experiments}\label{sec:experiments}

\subsection{Experimental Setup}\label{sec:setup}

The following hyper parameters (see Fig. \ref{fig:framework}) are used in our experiments.
\begin{itemize}
    \item Input filter window size to Hop-1: $I_0=2$,
    \item Input filter window size to Hop-2: $I_1=2$,
    \item Selected channel numbers in Hop-1 ($K_1$): $6\sim10$,
    \item Selected channel numbers in Hop-2 ($K_2$): $20\sim30$.
\end{itemize}
The window size of the analysis filter is the same as the generation window
size. All windows are non-overlapping with each other. The actual
channel numbers $K_1$ and $K_2$ are texture-dependent. That is, we
examine the energy distribution based on the PCA eigenvalues and
choose the knee point where the energy becomes flat.

\subsection{An Example: Brick Wall Generation}

\begin{figure*}[htb]
    \centering
    \subfloat[\centering Synthesized $32\times32$ Patches] 
    {{\includegraphics[width=0.45\linewidth]{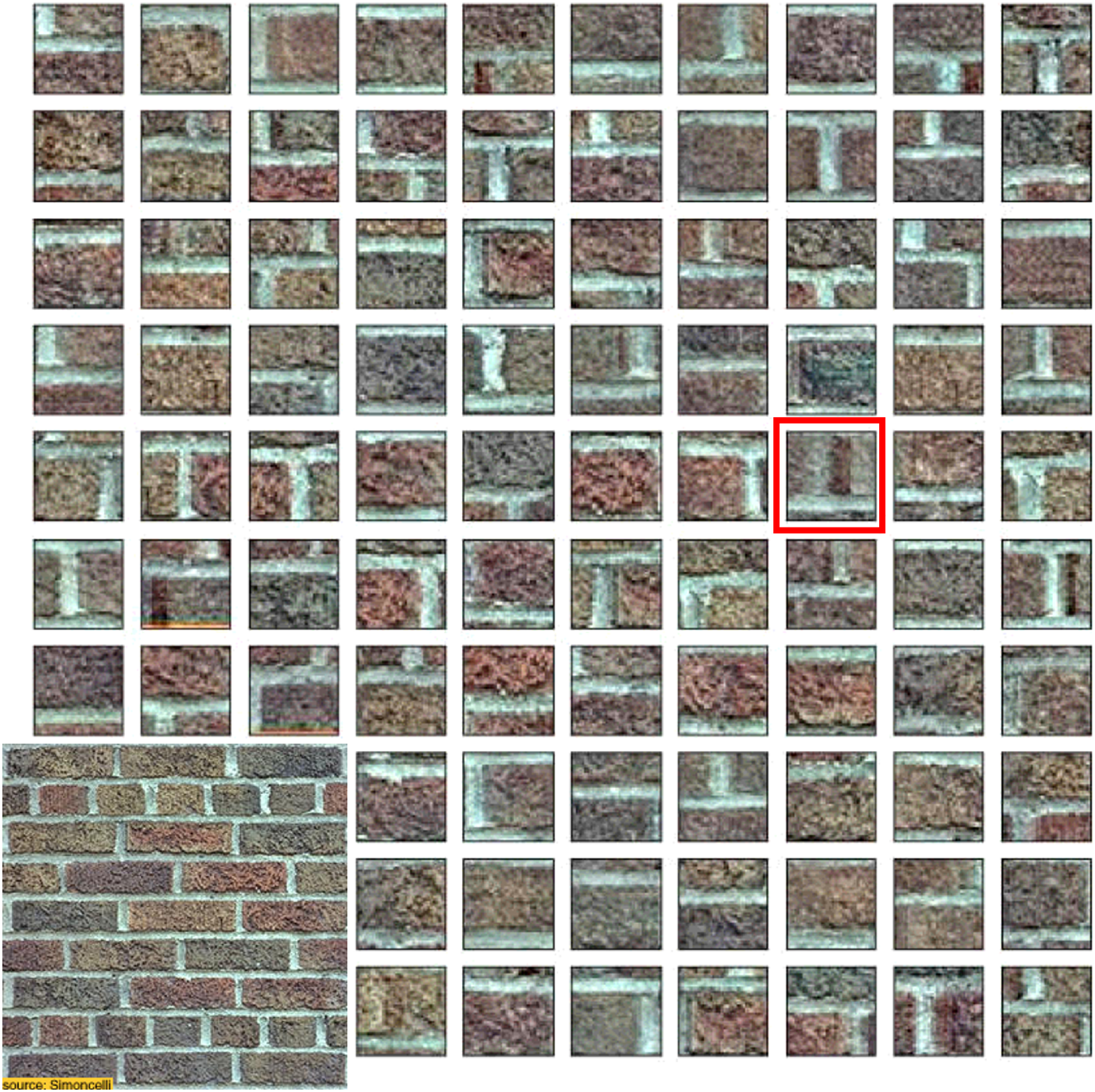} }}%
    \qquad
    \subfloat[\centering Stitched Images with $32\times32$ Patches] 
    {{\includegraphics[width=0.45\linewidth]{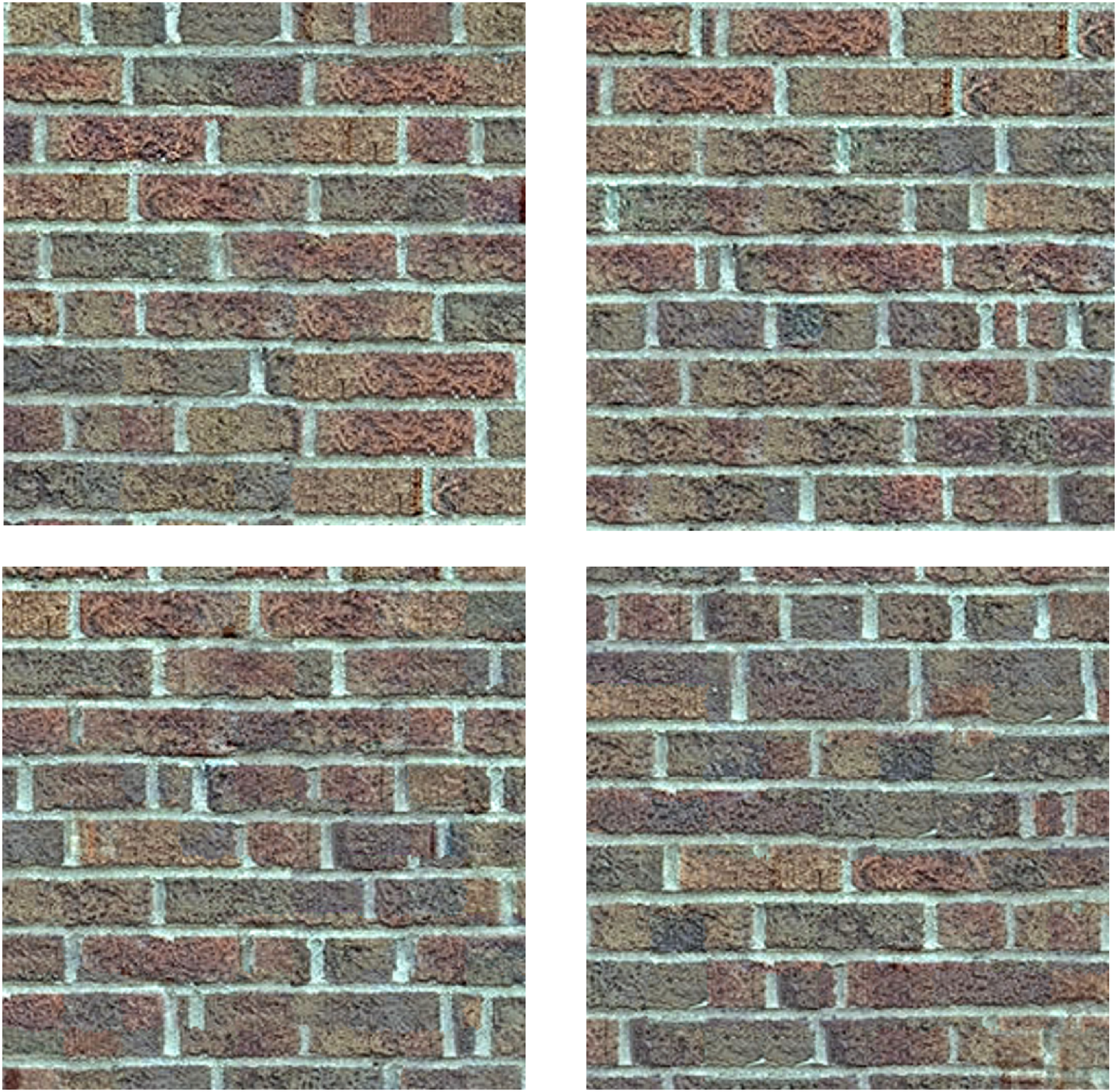} }}%

    \subfloat[\centering Synthesized $64\times64$ Patches] 
    {{\includegraphics[width=0.45\linewidth]{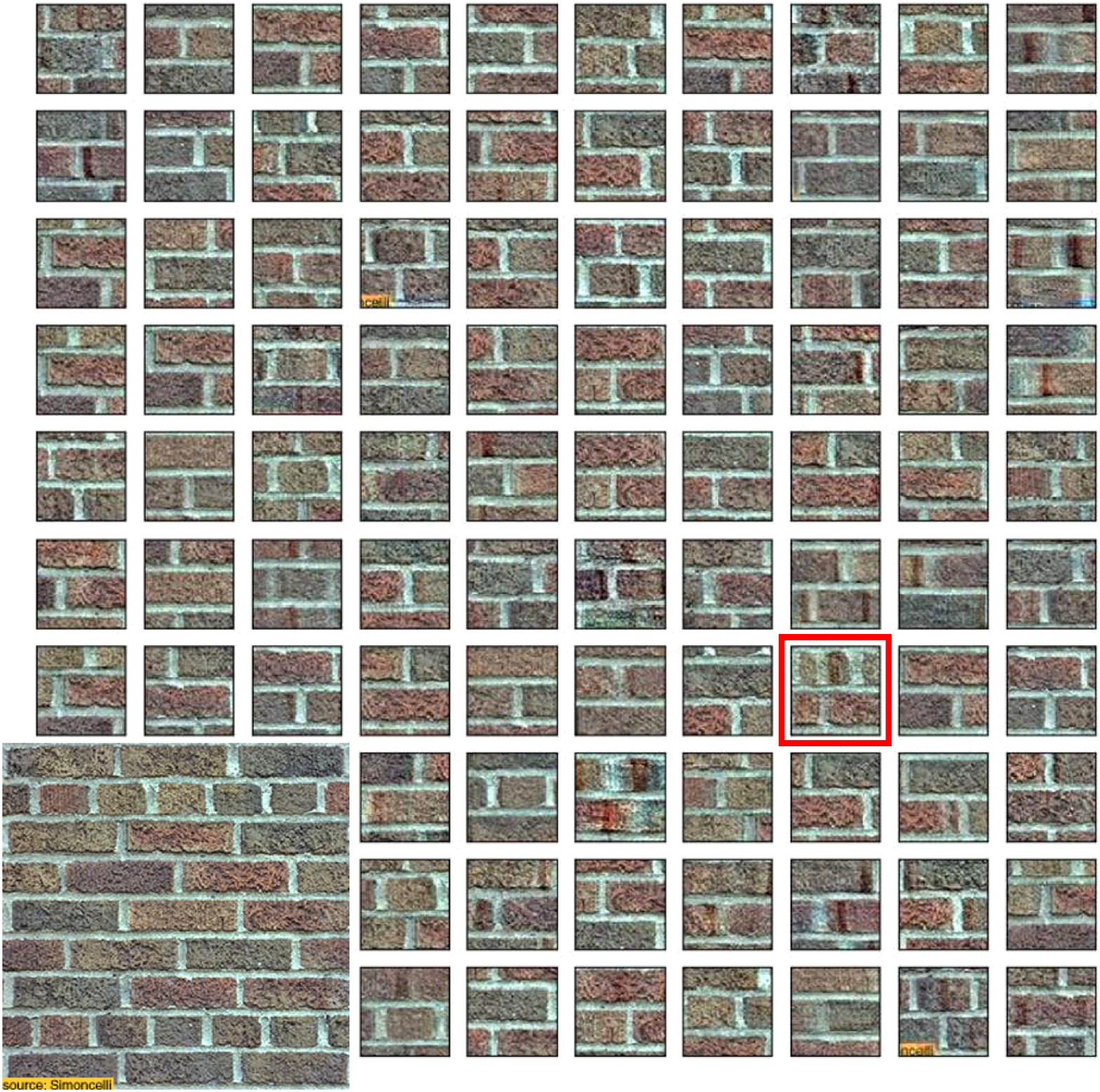} }}%
    \qquad
    \subfloat[\centering Stitched Images with $64\times64$ Patches] 
    {{\includegraphics[width=0.45\linewidth]{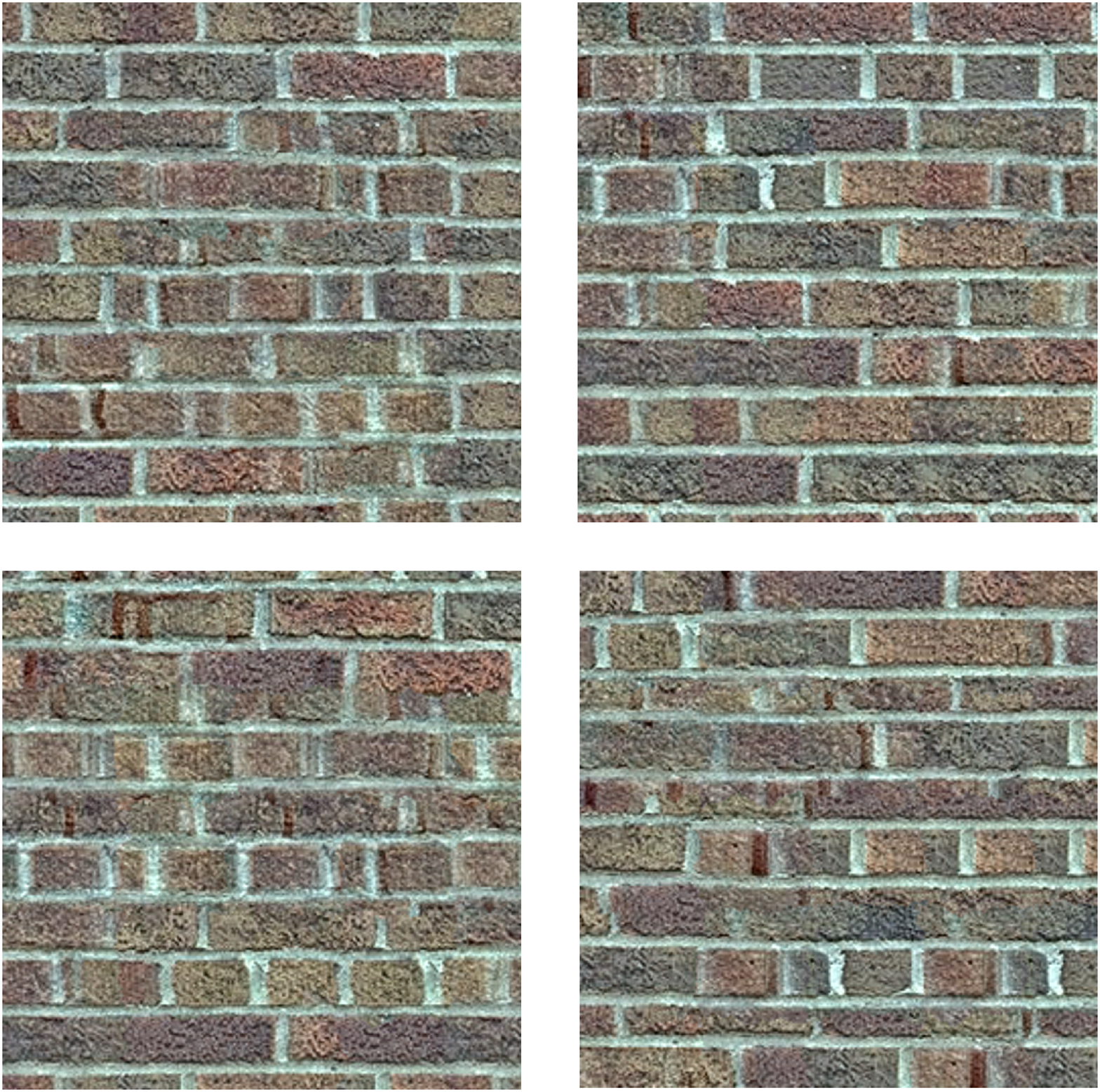} }}%
\caption{Examples of generated $brick\_wall$ texture patches and
stitched images of larger sizes, where the image in the bottom-left
corner is the exemplary texture image and the patches highlighted by red
squared boxes are unseen patterns.}\label{fig:synpatch}%
\end{figure*}

We show generated \emph{brick\_wall} patches of size $32\times32$ and
$64\times64$ in Figs.~\ref{fig:synpatch}(a) and (c).  We performed
two-stage c/w Saab transforms on $32\times32$ patches and three-stage
c/w Saab transforms on $64\times64$ patches, whose core subspace
dimensions are 1728 and 4032, respectively. Patches in these figures
were synthesized by running the TGHop method in one hundred rounds.
Randomness in each round primarily comes from two factors: 1) random
cluster selection, and 2) random seed vector generation. 

Generated patches retain the basic shape of bricks and the diversity of
brick texture. We observe some unseen patterns generated by TGHop, which
are highlighted by red squared boxes in Fig.~\ref{fig:synpatch} (a) and
(c). As compared with generated $32\times32$ patches, generated
$64\times64$ patches were sometimes blurry (e.g., the one in the upper
right corner) due to a higher source dimension. 

\begin{figure*}[htb]
    \centering
    \subfloat[\centering $3072\rightarrow2560\rightarrow2048$] 
    {{\includegraphics[width=0.2\linewidth]{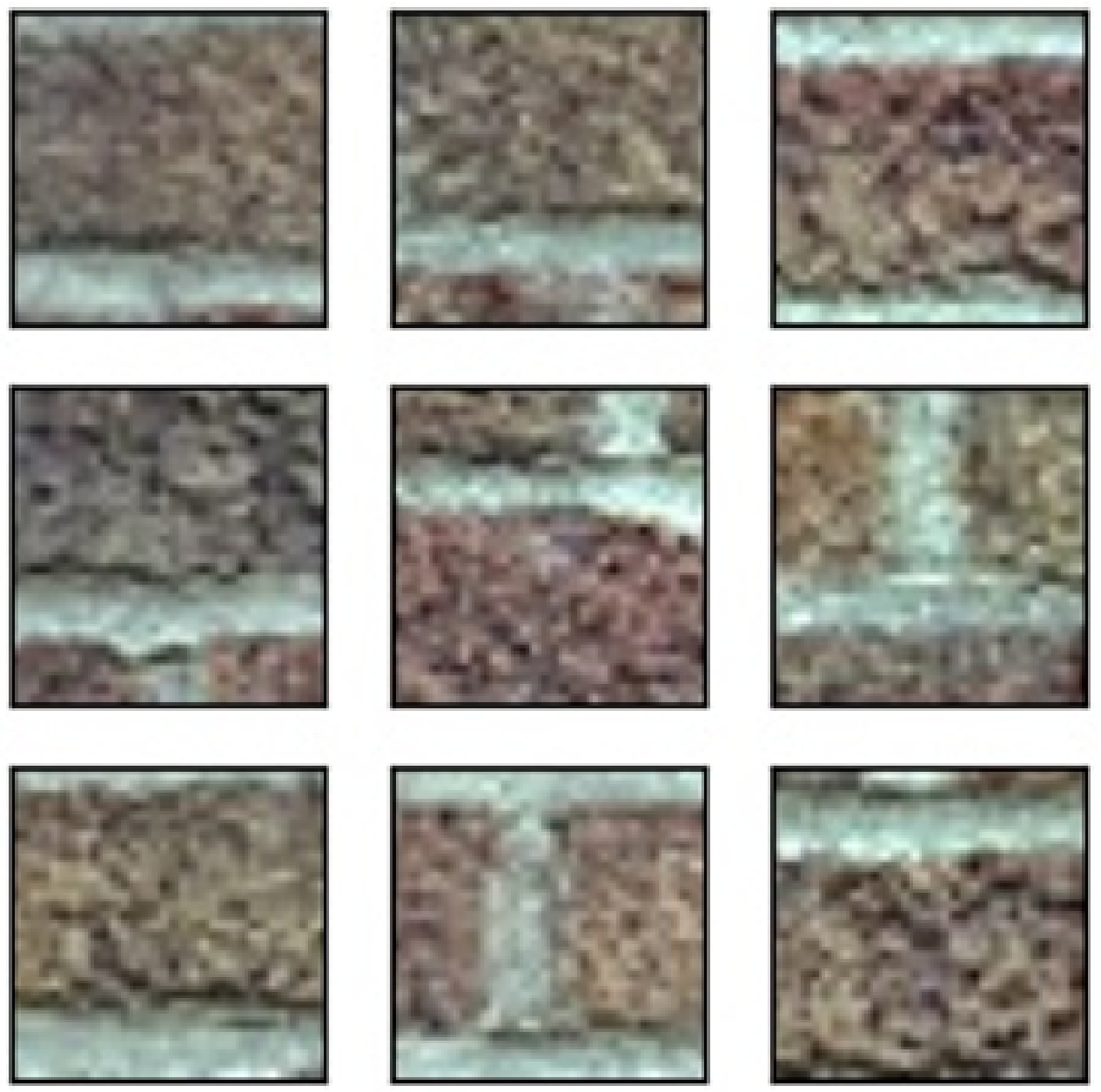} }}%
    \qquad
    \subfloat[\centering $3072\rightarrow1536\rightarrow768$] 
    {{\includegraphics[width=0.2\linewidth]{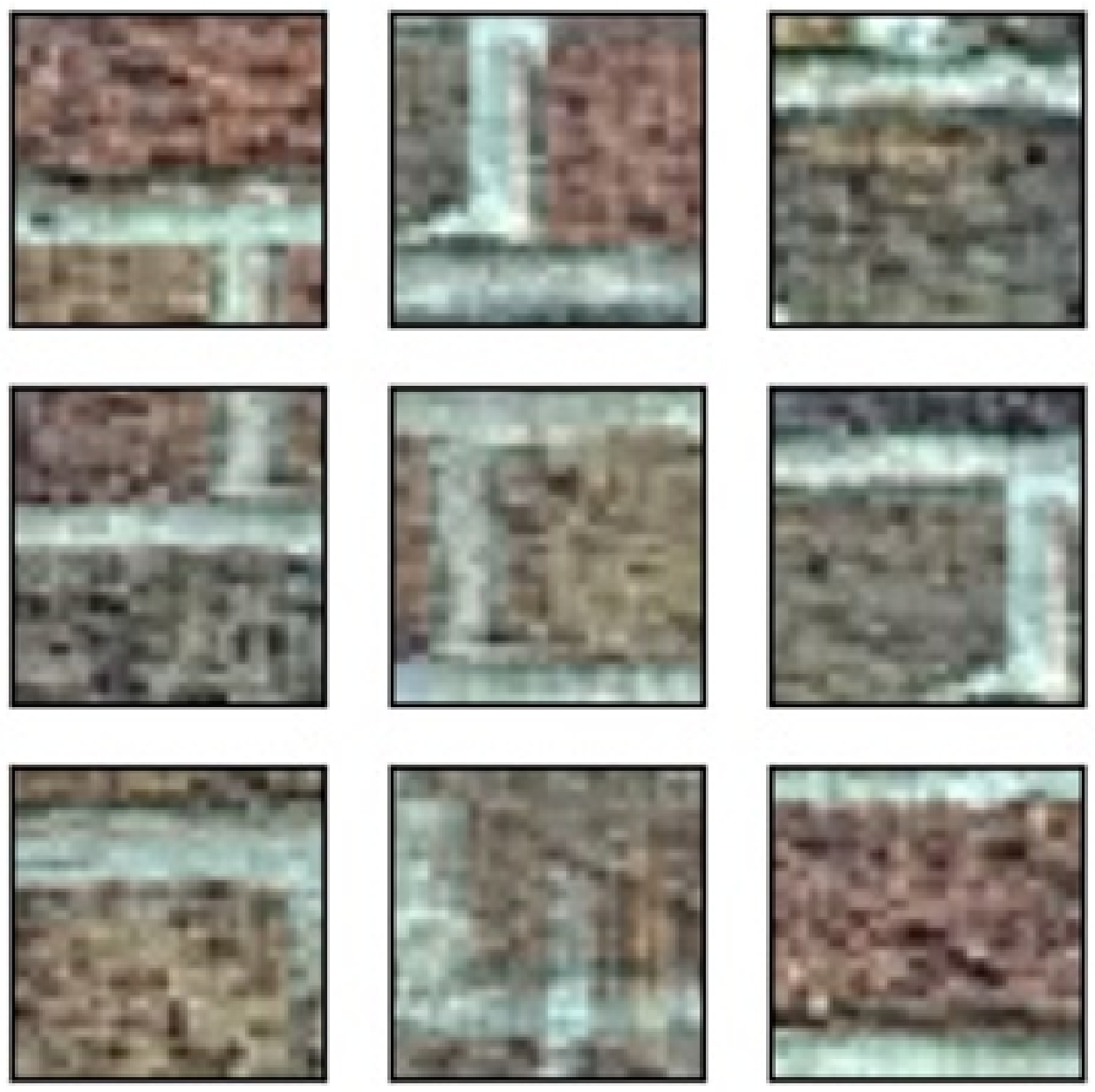} }}%
    \qquad
    \subfloat[\centering $3072\rightarrow1280\rightarrow512$] 
    {{\includegraphics[width=0.2\linewidth]{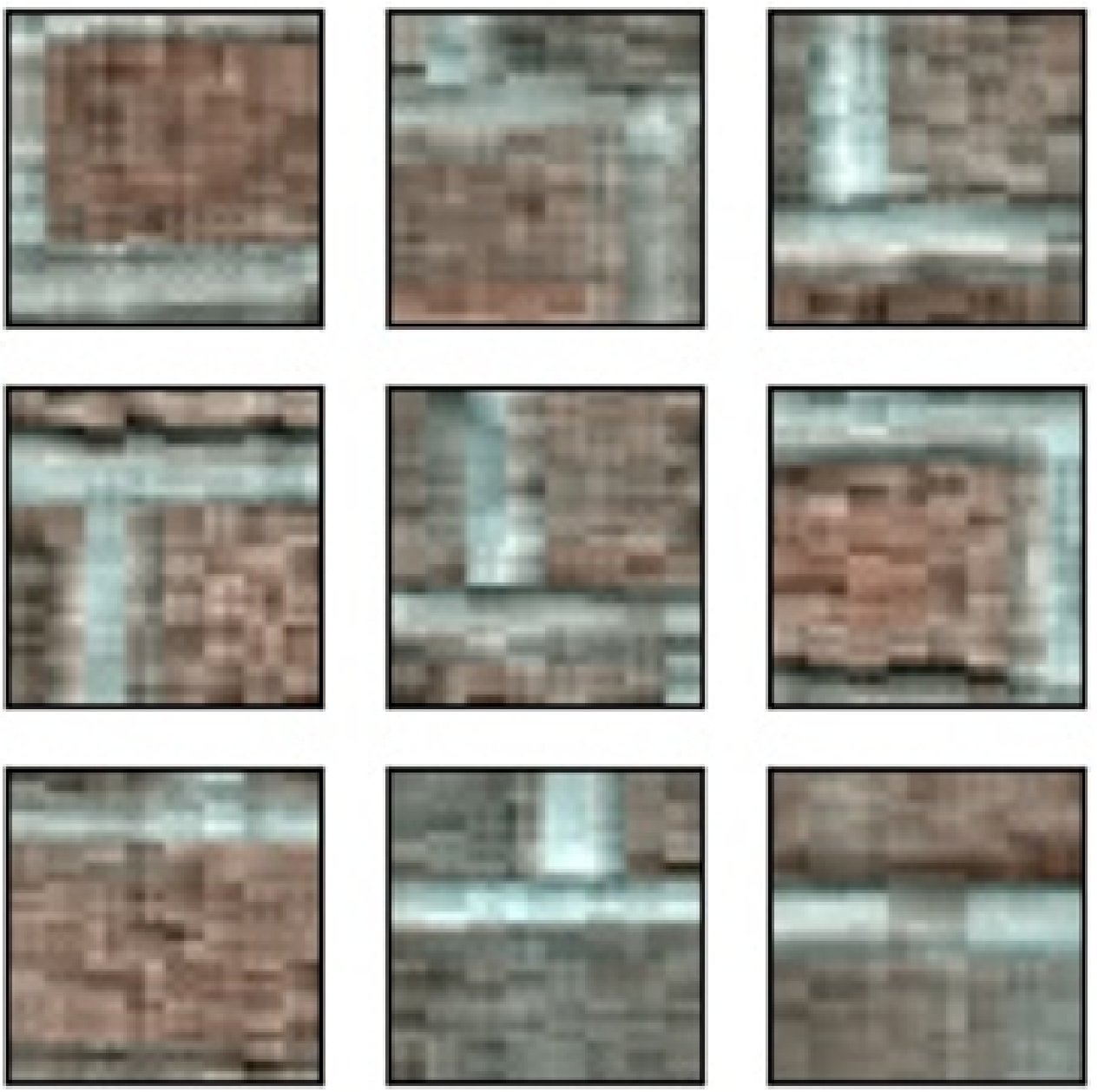} }}%
    \qquad
    \subfloat[\centering $3072\rightarrow768\rightarrow192$] 
    {{\includegraphics[width=0.2\linewidth]{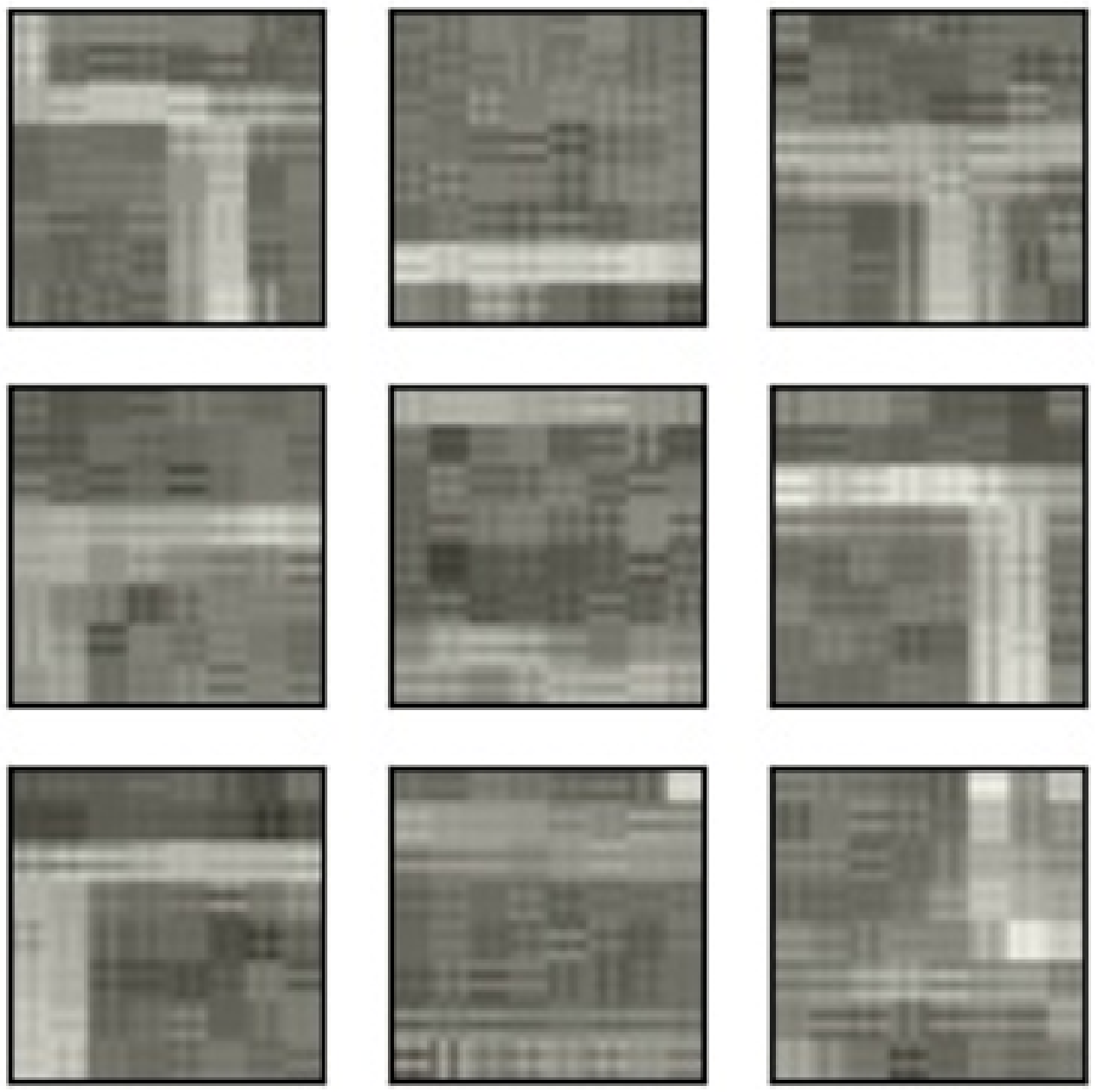} }}%
\caption{Generated patches using different settings, where the numbers below
the figure indicates the dimensions of $S_0$, $S_1$ and $S_2$,
respectively.} \label{fig:dimanalysis}%
\end{figure*}

As a non-parametric model, TGHop can choose multiple settings under the
same pipeline. For example, it can select different channel numbers in
$\tilde{S}_1$ and $\tilde{S}_2$ to derive different generation results.
Four settings are listed in Table~\ref{table:settings}. The
corresponding generation results are shown in
Fig.~\ref{fig:dimanalysis}. Dimensions decrease faster from (a) to (d).
The quality of generated results becomes poorer due to smaller
dimensions of the core subspace, $\tilde{S}_2$, and the intermediate
subspace, $\tilde{S}_1$. 

\begin{table}[htb]
\begin{center}
\caption{The settings of four generation models.} \label{table:settings}
\begin{tabular}{cccc} \hline
Setting & $\tilde{D}_0$ & $\tilde{D}_1$ & $\tilde{D}_2$ \\ \hline
a       & 3072          & 2560          & 2048 \\ 
b       & 3072          & 1536          & 768 \\ 
c       & 3072          & 1280          & 512 \\ 
d       & 3072          & 768           & 192 \\ \hline
\end{tabular}
\end{center}
\end{table}

To generate larger texture images, we first generate 5,000 texture
patches and perform image quilting~\cite{efros2001image} with them. The
results after quilting are shown in Figs.~\ref{fig:synpatch} (b) and
(d).  All eight images are of the same size, i.e., $256\times256$. They are
obtained using different initial patches in the image quilting process.
By comparing the two sets of stitched images, the global structure of
the brick wall is better preserved using larger patches (i.e. of size
$64 \times 64$) while its local detail is a little bit blurry sometimes. 

\subsection{Performance Benchmarking with DL-based Methods}

\subsubsection{Visual Quality Comparison} 

\begin{figure*}[!htbp]
\begin{center}
\includegraphics[width=\textwidth, height=0.92\textheight, keepaspectratio]{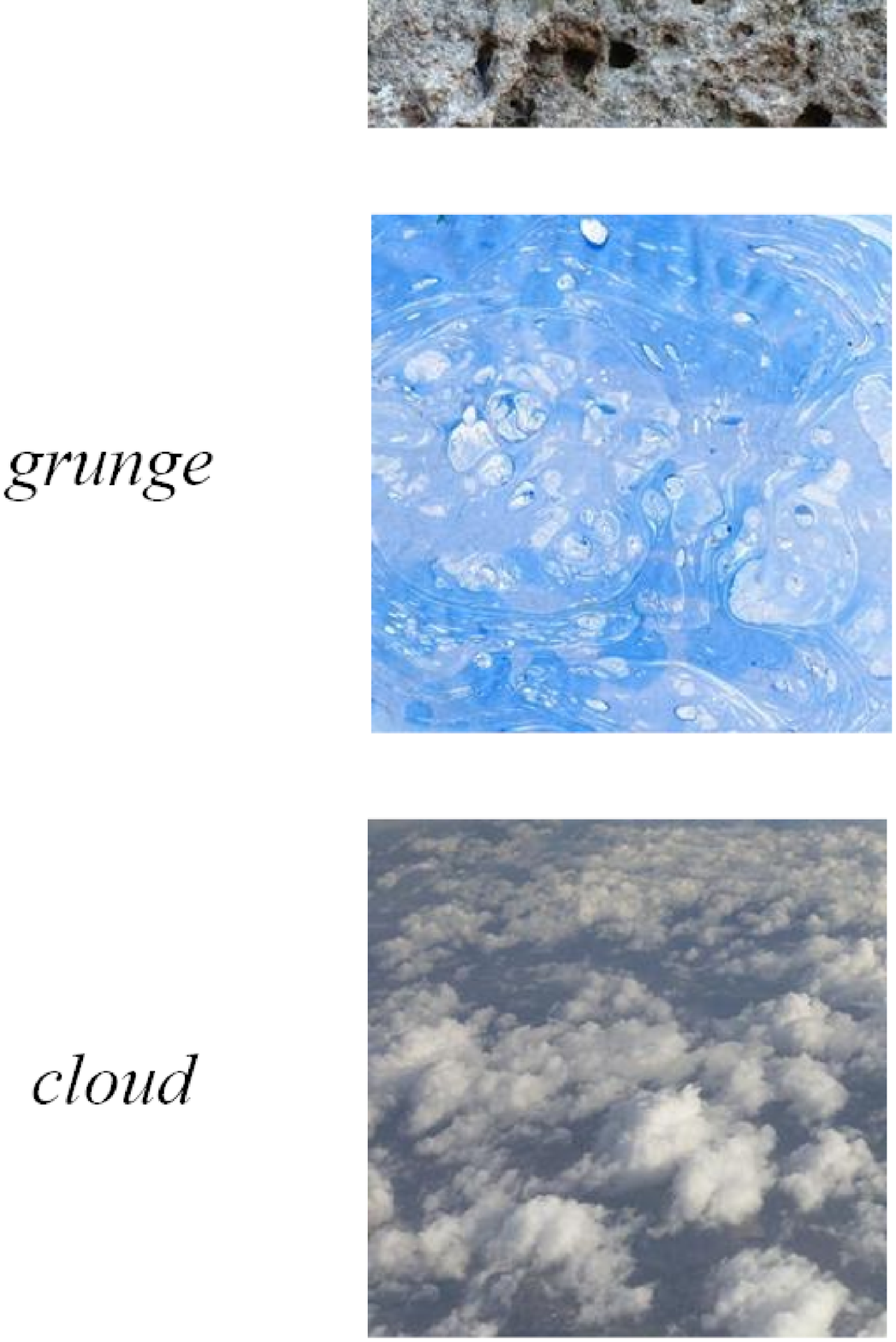}
\end{center}
\caption{Comparison of texture images generated by two DL-based methods
and TGHop (from left to right): exemplary texture images, texture images
generated by \cite{gatys2015texture}, by \cite{ustyuzhaninov2017does},
two examples by TGHop without spatial dimension reduction (SDR) and one
example by TGHop with SDR.} \label{fig:synimg}
\end{figure*}

\begin{figure*}[!htbp]
\begin{center}
\includegraphics[width=0.9\textwidth, height=\textheight, keepaspectratio]{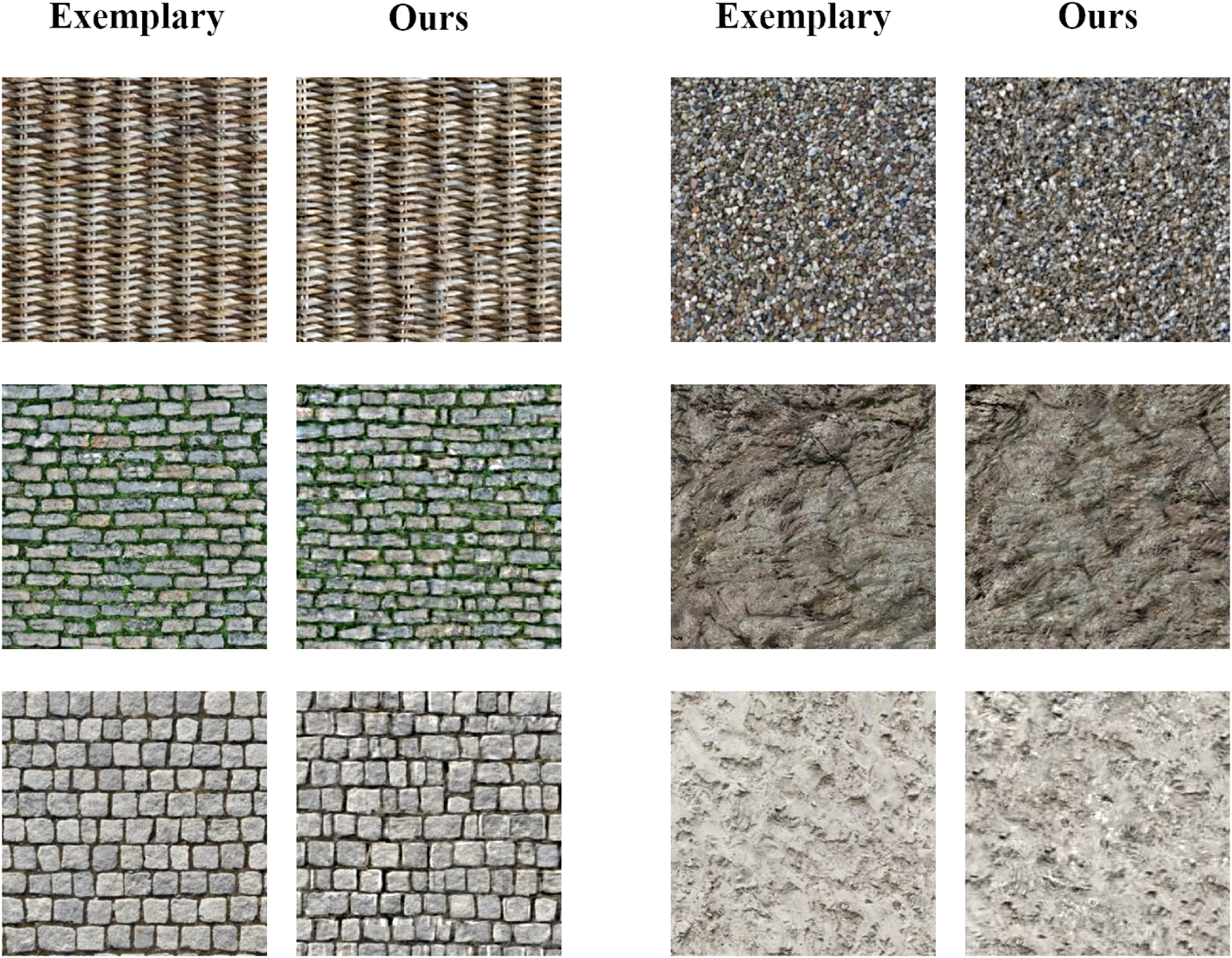}
\end{center}
\caption{More texture images generated by TGHop.} \label{fig:imgcollection}
\end{figure*}

The quality of generated texture is usually evaluated by human eyes. A
diversity loss function was proposed to measure texture diversity for
DL-based methods~\cite{shi2020fast, li2017diversified}. Since TGHop dose
not have a loss function, we show generated results of two DL-based
methods and TGHop side by side in Fig.~\ref{fig:synimg} for 10 input
texture images collected from~\cite{gatys2015texture,ustyuzhaninov2017does, portilla2000parametric} or the Internet. The benchmarking DL methods were proposed by Gatys {\em et al.}~\cite{gatys2015texture} and Ustyuzhaninov {\em et al.}
\cite{ustyuzhaninov2017does}. By running their codes, we show their
results in the second and third columns of Fig.~\ref{fig:synimg},
respectively, for comparison. These results are obtained by default
iteration numbers; namely, 2000 in~\cite{gatys2015texture} and 4000
in~\cite{gatys2015texture}. The results of TGHop are shown in the last
three columns. The left two columns are obtained without spatial
dimension reduction (SDR) in two different runs while the last column is
obtained with SDR. There is little quality degradation after dimension
reduction of $S_2$ with SDR.  For \emph{meshed} and \emph{cloud}
textures, the brown fog artifact in ~\cite{gatys2015texture,ustyuzhaninov2017does} is apparent. In contrast, it does not exist in TGHop. More generated images using TGHop are given in Fig.~\ref{fig:imgcollection}. As shown in Figs. ~\ref{fig:synimg}
and~\ref{fig:imgcollection}, TGHop can generate high quality and visually pleasant texture images. 

\subsubsection{Comparison of Generation Time} 

We compare the generation time of different methods in
Table~\ref{table:time}. All experiments were conducted on the same
machine composed of 12 CPUs (Intel Core i7-5930K CPU at 3.50GHz) and 1
GPU (GeForce GTX TITAN X). GPU was needed in two DL-based methods but
not in TGHop. We set the iteration number to 1000
for~\cite{gatys2015texture} and 100 for~\cite{ustyuzhaninov2017does}.
TGHop generated 10K $32\times32$ patches for texture quilting.  For all
three methods, we show the time needed in generating one image of size
$256\times 256$ in Table~\ref{table:time}, TGHop generates one texture
image in 291.25 seconds while Gatys' method and Ustyuzhaninov's method
demand 513.98 and 949.64 seconds, respectively.  TGHop is significantly
faster. 

\begin{table}[htb]
\begin{center}
\caption{Comparison of time needed to generate one texture image.} \label{table:time}
\begin{tabular}{ccc} \\ \hline
Methods                                           &  Time (seconds) & Factor \\ \hline
Ustyuzhaninov et al.~\cite{ustyuzhaninov2017does} &  949.64  &  4.62x    \\
Gatys et al.~\cite{gatys2015texture}              &  513.98  &  2.50x    \\ 
TGHop with analysis overhead                      &  291.25  &  1.42x    \\ 
TGHop w/o analysis overhead                       &  205.50  &  1x   \\ \hline
\end{tabular}
\end{center}
\end{table}

We break down the generation time of TGHop into three parts: 1)
successive subspace analysis (i.e., the forward path), 2) core and
successive subspace generation (i.e., the reverse path) and 3) the
quilting process. The time required for each part is shown in
Table~\ref{table:processtime}. They demand 85.75, 197.42 and 8.08
seconds, respectively. To generate multiple images from the same
exemplary texture, we run the first part only once, which will be shared
by all generated texture images, and the second and third parts multiple
times (i.e., one run for a new image). As a result, we can view the
first part as a common overhead and count the last two parts as the time
for single texture image generation. This is equal to 205.5 seconds.
The two DL benchmarks do not have such a breakdown and need to go
through the whole pipeline to generate one new texture image. 

\begin{table}[htb]
\begin{center}
\caption{The time of three processes in our method.}\label{table:processtime}
\begin{tabular}{cc} \\ \hline
Processes                &   Time (seconds)  \\ \hline
Analysis (Forward Path)  &  85.75 \\
Generation (Reverse Path)&  197.42 \\
Quilting                 &    8.08 \\  \hline
\end{tabular}
\end{center}
\end{table}

\subsection{Comparison of Model Sizes}

The model size is measured by the number of parameters. The size of
TGHop is calculated below.  
\begin{itemize}
\item Two-stage c/w Saab Transforms \\
The forward analysis path and the reverse generation path share the same
two-stage c/w Saab transforms. For an input RGB patch, the input tensor
of size $2\times 2 \times 3=12$ is transformed into a $K_1$-D tensor in
the first-stage transform, leading a filter size of $12 K_1$ plus one
shared bias. For each of $K_1$ channels, the input tensor of size
$2\times 2$ is transformed into a $K_2$-D tensor in the second stage
transform. The total parameter number for all $K_1$ channels is $4 K_1
K_2$ plus $K_1$ biases.  Thus, the total number of parameters in the
two-stage transforms is $13K_1+4K_1K_2+1$. 
\item Core Sample Generation \\ 
Sample generation in the core contains two modules: spatial dimension
reduction (SDR) and independent components histogram matching (ICHM).
For the first module, SDR is implemented by $K_2$ PCA transforms, where
the input of size $8\times 8=64$ and the output is a $K_{r_i}$
dimensional vector, yielding the size of each PCA transformation matrix
to be $64\times K_{r_i}$. The total number of parameters is $64\times
\sum_{i=1}^{K_2}{K_{r_i}}=64D_r$, where $D_r$ is the dimension of the
concatenated output vector after SDR. 
For the second module, it has three components:
\begin{enumerate}
\item Interval representation $p_0,\dots,p_{N-1}$ \\
$N$ parameters are needed for each cluster.
\item Transform matrices of FastICA \\
If the input vector is of dimension $D_r$ and the output dimension of
FastICA is $K_{c_1},\dots,K_{c_{i}}$ for the $i$th cluster, $i=1, \cdots, W$, 
the total parameter number of all transforms matrices is $D_r F$, where
$F=\sum_{i=1}^{N}{K_{c_i}}$ is the number of CDFs. 
\item Codebook size of quantized CDFs  \\
The codebook contains the index, the maximum and the minimum values for
each CDF. Furthermore, we have $W$ clusters of CDF, where all CDFs in 
each cluster share the same bin structure of 256 bins. As a result, 
the total parameter number is $3F+256W$. 
\end{enumerate}
By adding all of the above together, the total parameter number in
core sample generation is $64D_r+N+(D_r+3)F+256W$. 
\end{itemize}

\begin{figure*}[htb]
    \centering
    \subfloat[\centering 50 clusters] {{\includegraphics[width=0.21\linewidth]{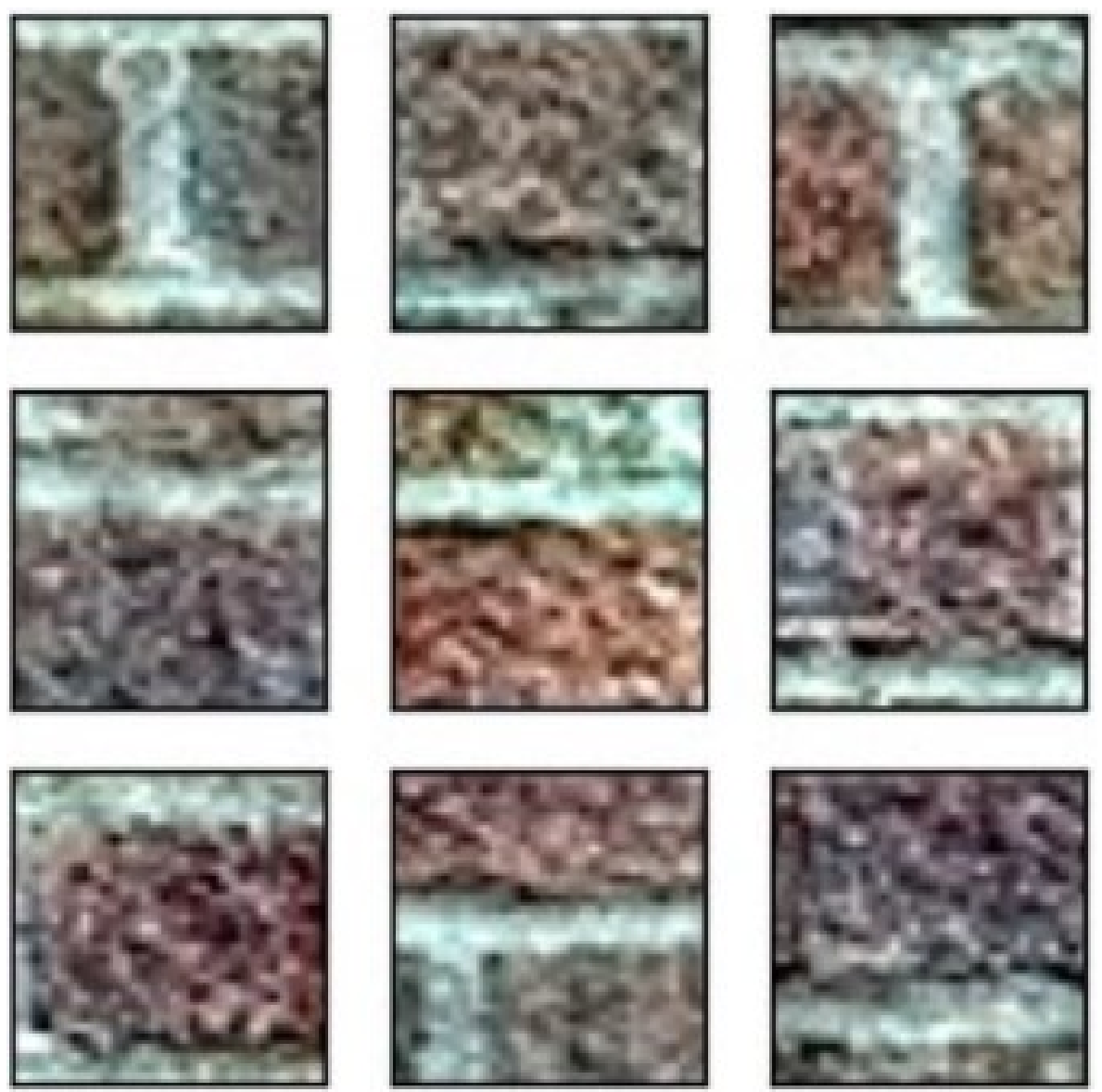} }}%
    \qquad
    \subfloat[\centering 80 clusters] {{\includegraphics[width=0.21\linewidth]{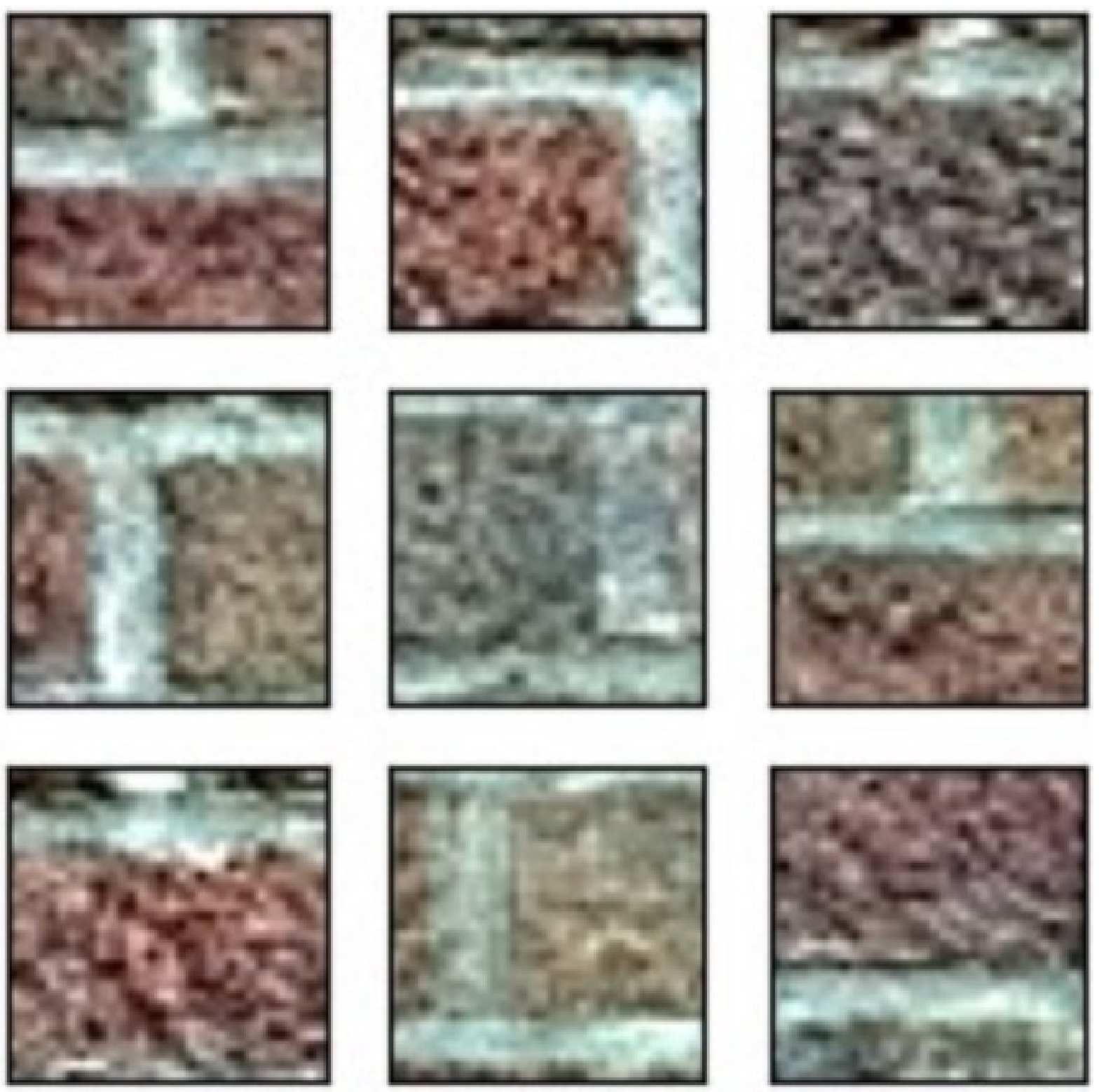} }}%
    \qquad
    \subfloat[\centering 110 clusters] {{\includegraphics[width=0.21\linewidth]{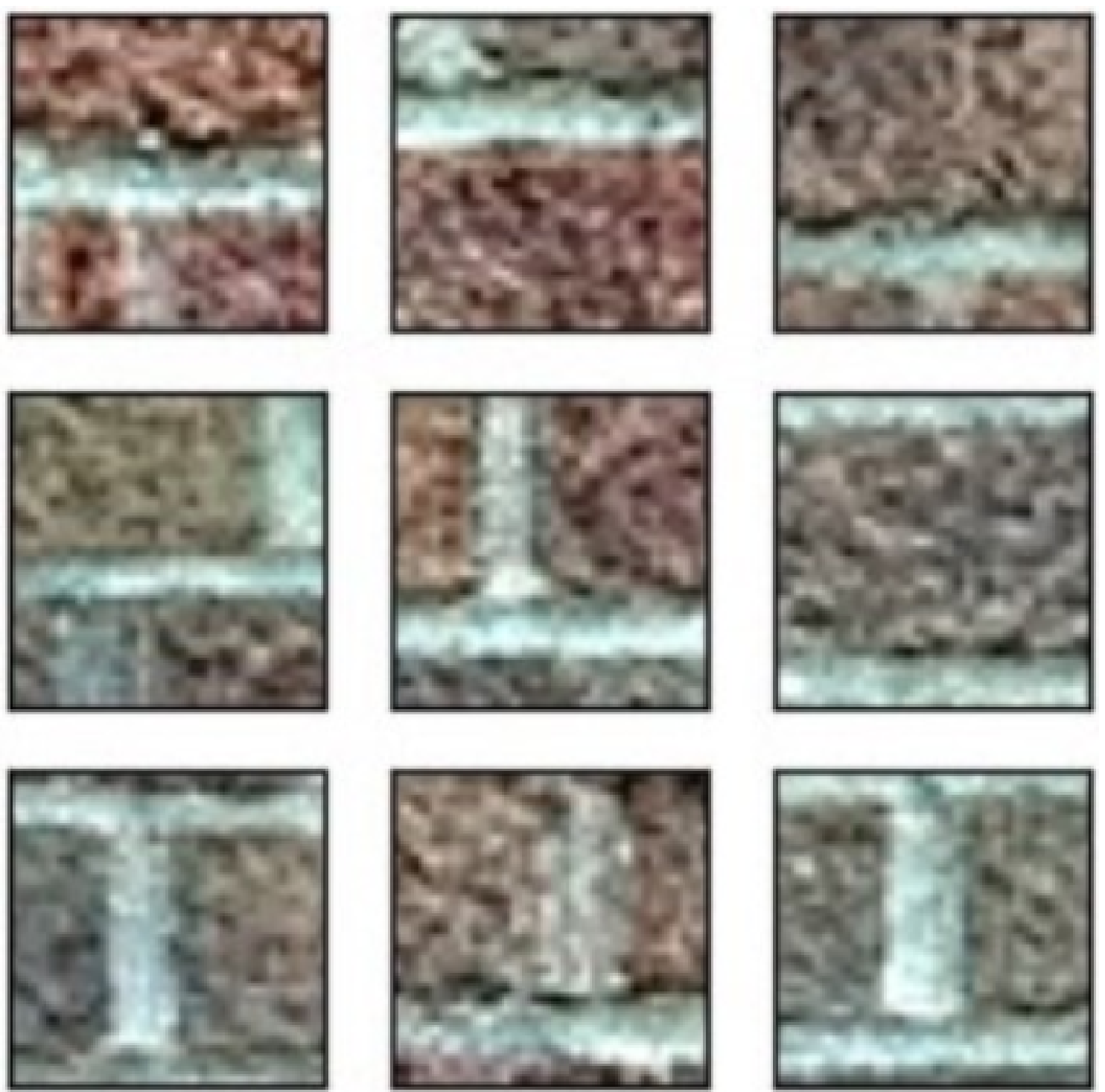} }}%
    \qquad
    \subfloat[\centering 200 clusters] {{\includegraphics[width=0.21\linewidth]{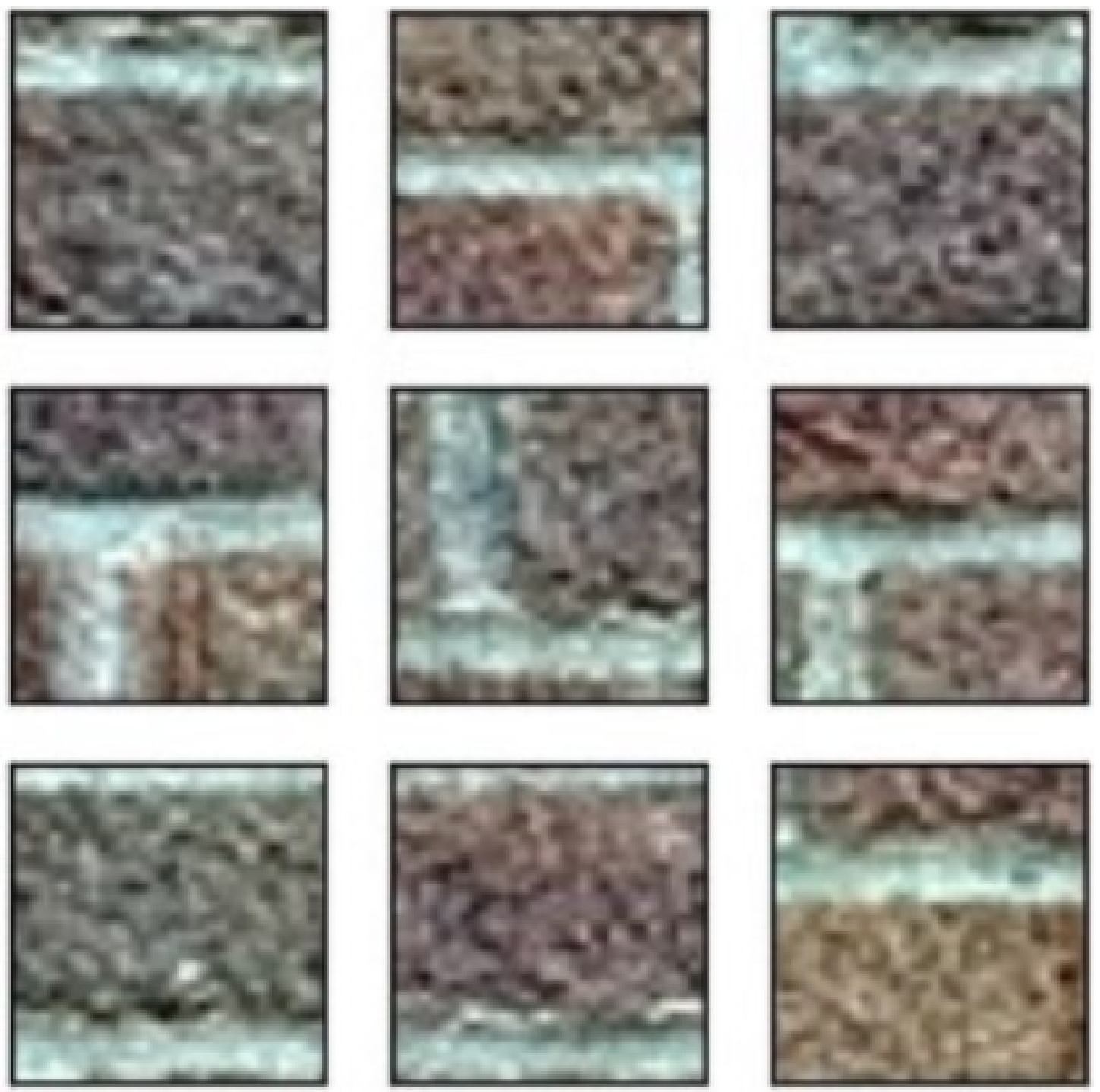} }}%
    \caption{Generated \emph{brick\_wall} patches using different cluster numbers in 
    independent component histogram matching.}\label{fig:numclu}%
\end{figure*}

\begin{figure*}[htb]%
    \centering
    \subfloat[\centering $\gamma=0$] {{\includegraphics[width=0.21\linewidth]{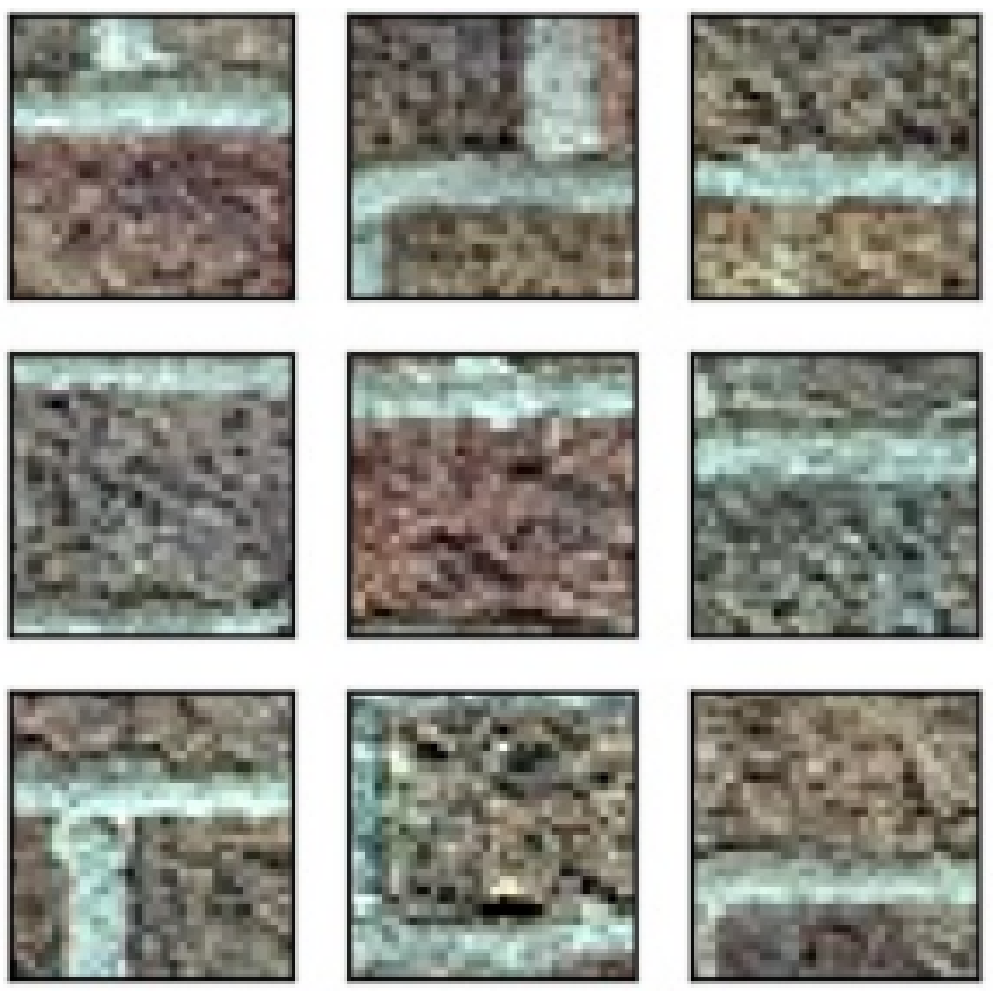} }}%
    \qquad
    \subfloat[\centering $\gamma=0.0005$] {{\includegraphics[width=0.21\linewidth]{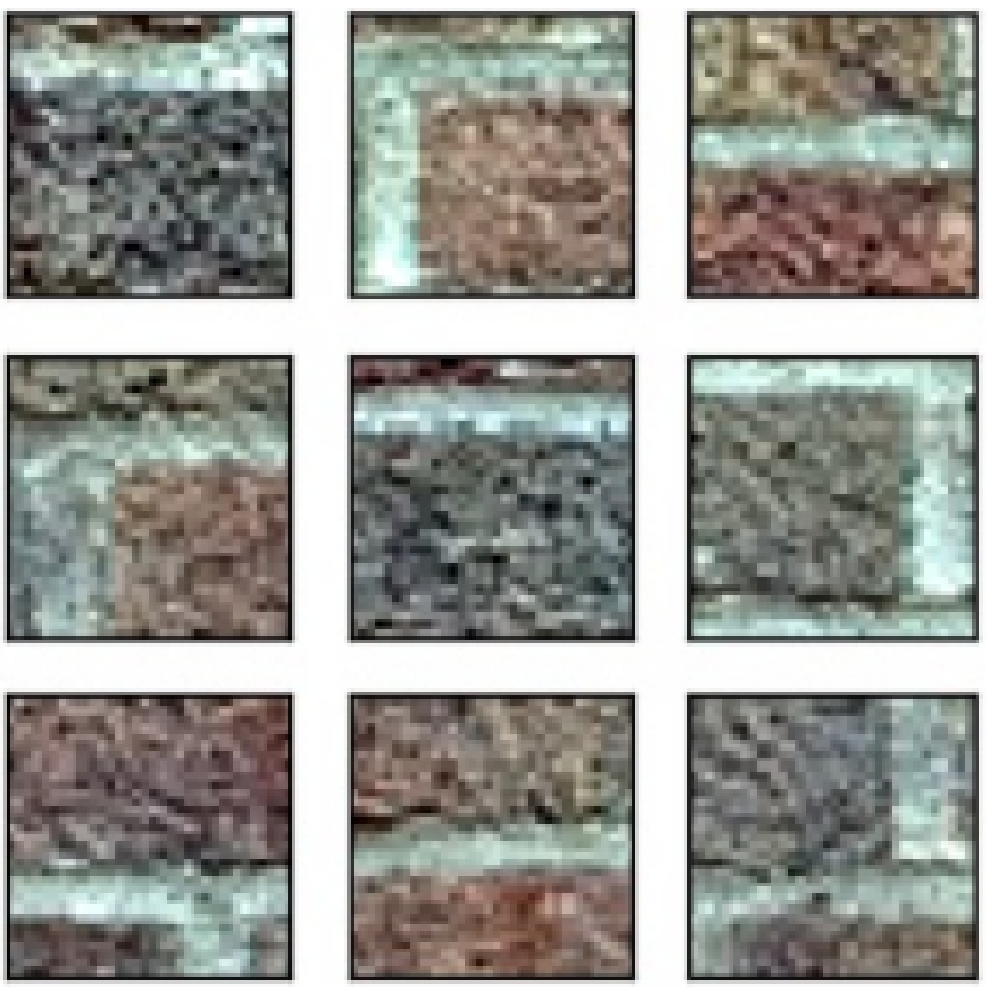} }}%
    \qquad
    \subfloat[\centering $\gamma=0.005$] {{\includegraphics[width=0.21\linewidth]{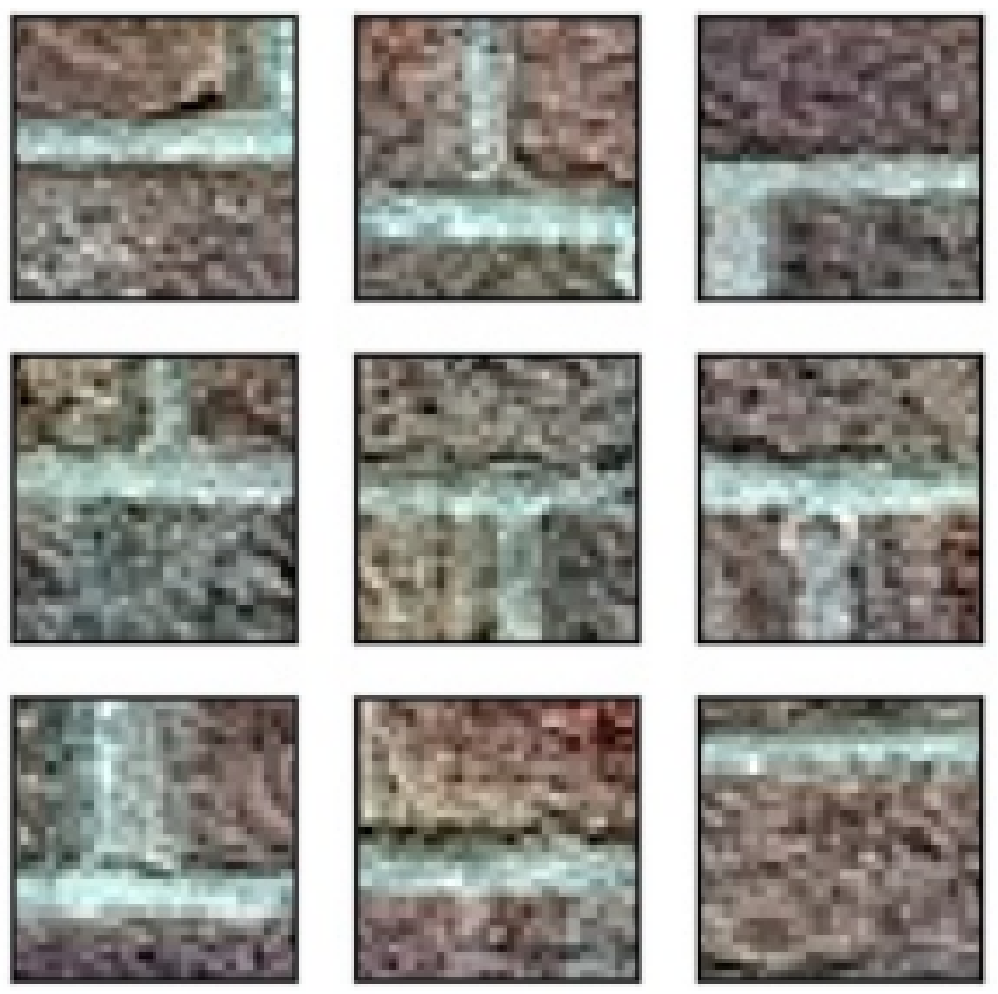} }}%

    \subfloat[\centering $\gamma=0.01$] {{\includegraphics[width=0.21\linewidth]{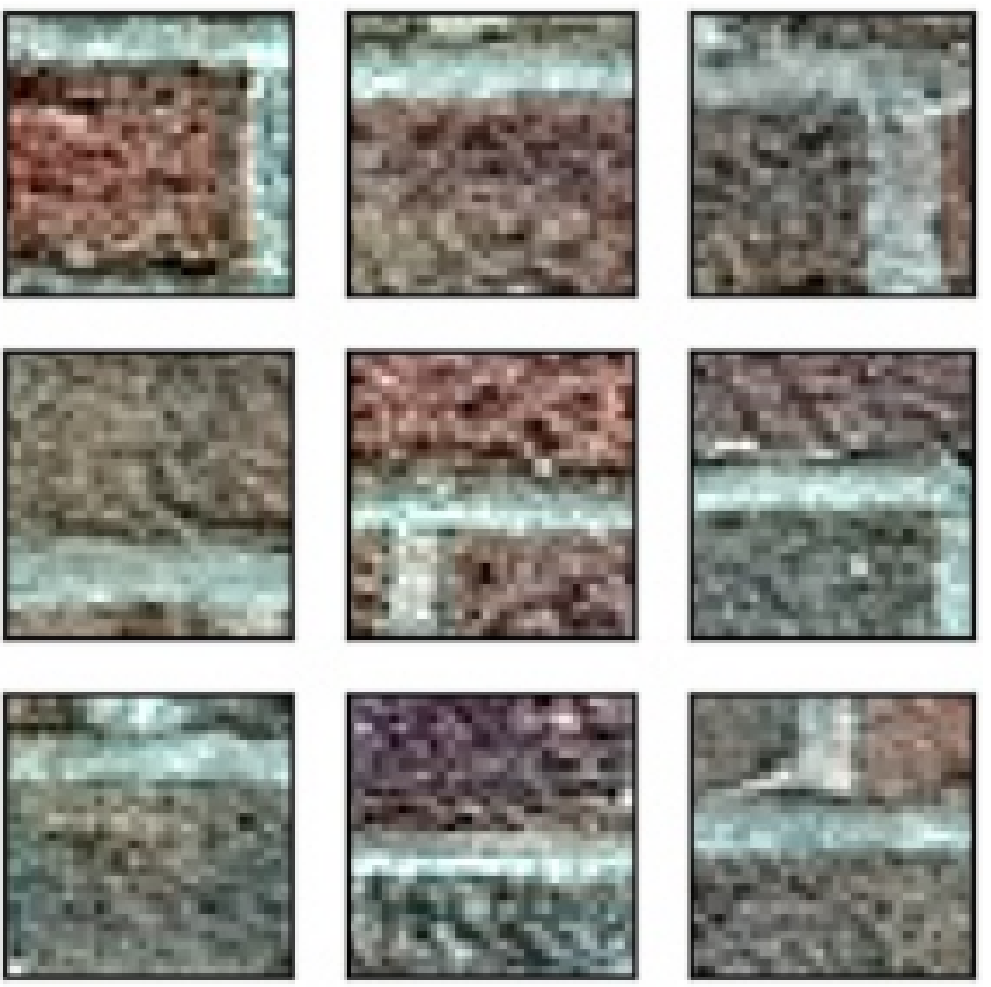} }}%
    \qquad
    \subfloat[\centering $\gamma=0.02$] {{\includegraphics[width=0.21\linewidth]{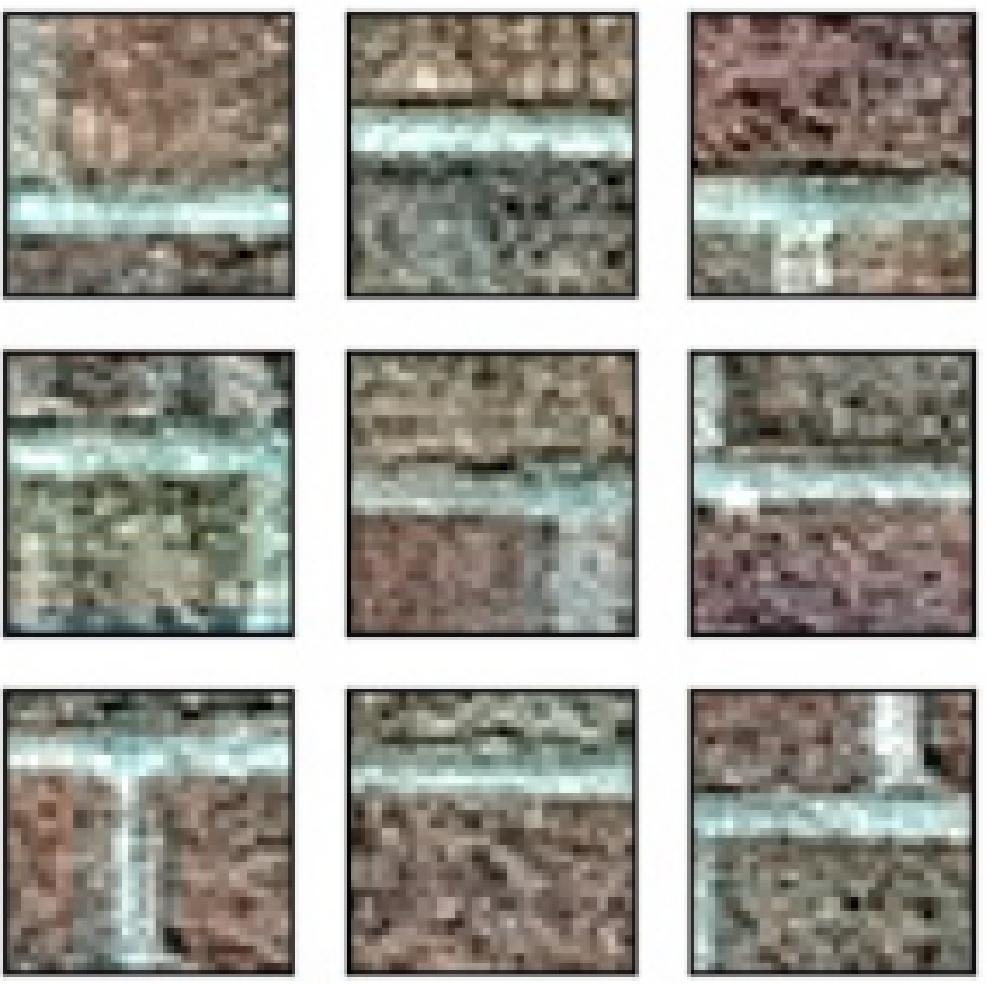} }}%
    \qquad
    \subfloat[\centering $\gamma=0.03$] {{\includegraphics[width=0.21\linewidth]{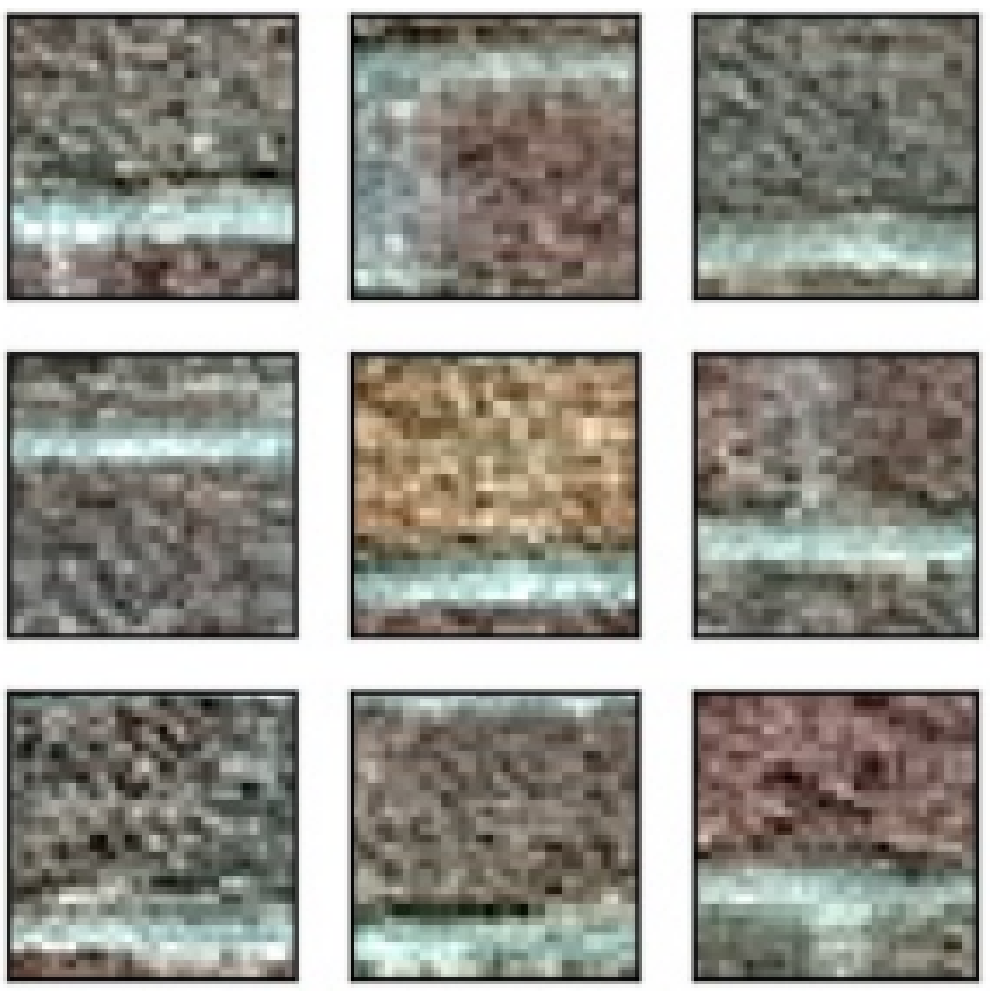} }}%

    \subfloat[\centering $\gamma=0.04$] {{\includegraphics[width=0.21\linewidth]{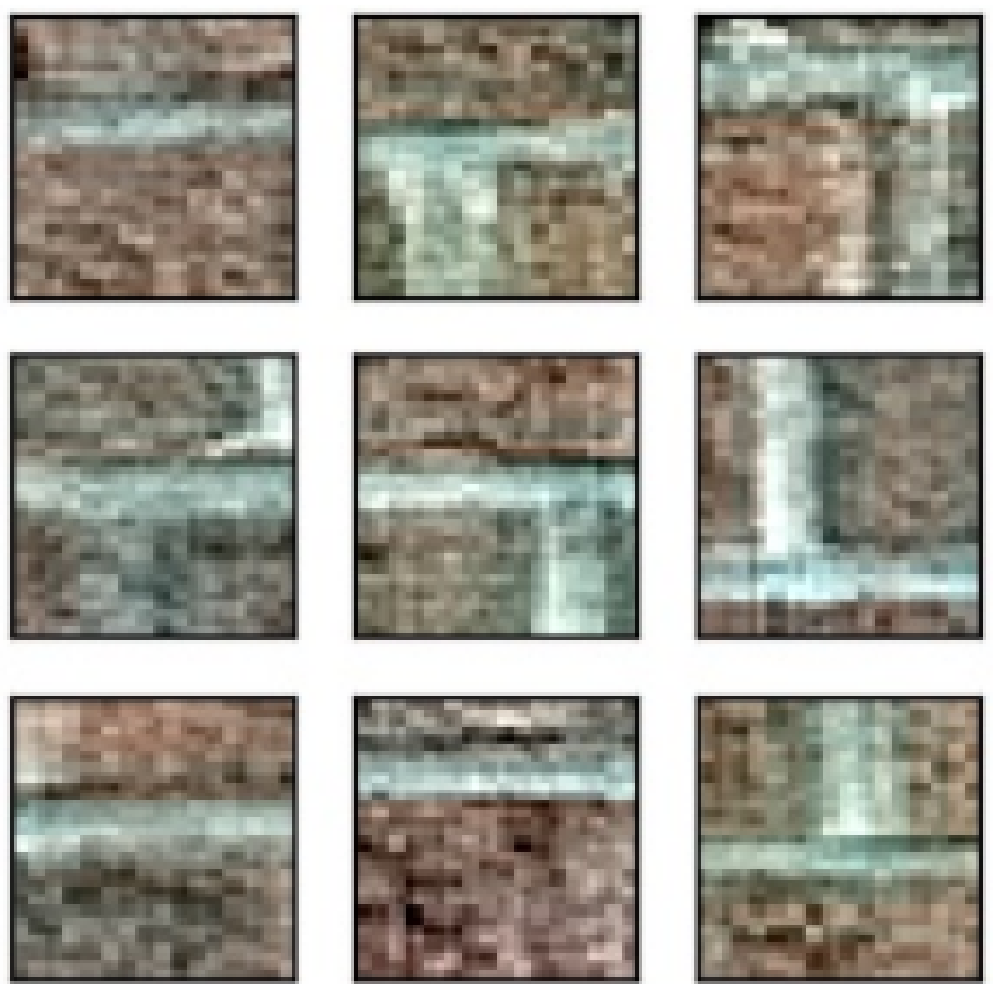} }}%
    \qquad
    \subfloat[\centering $\gamma=0.05$] {{\includegraphics[width=0.21\linewidth]{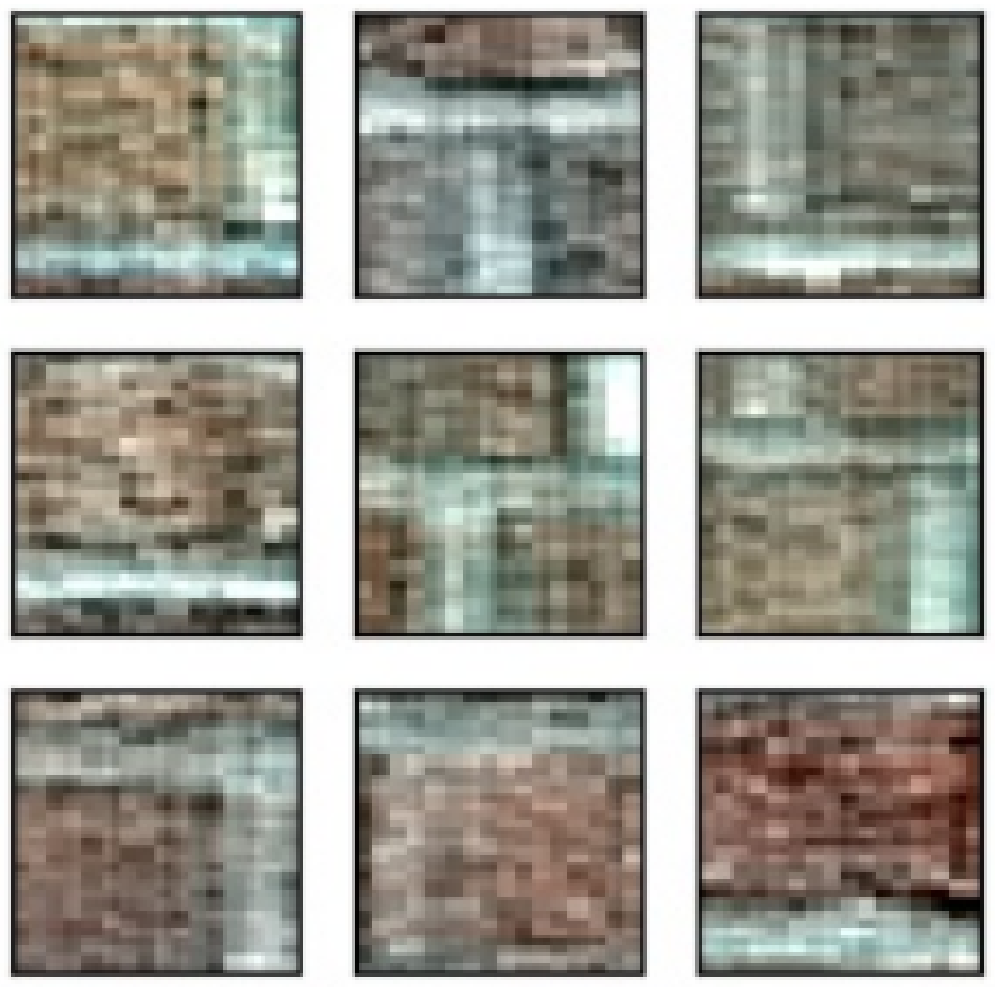} }}%
    \qquad
    \subfloat[\centering $\gamma=0.1$] {{\includegraphics[width=0.21\linewidth]{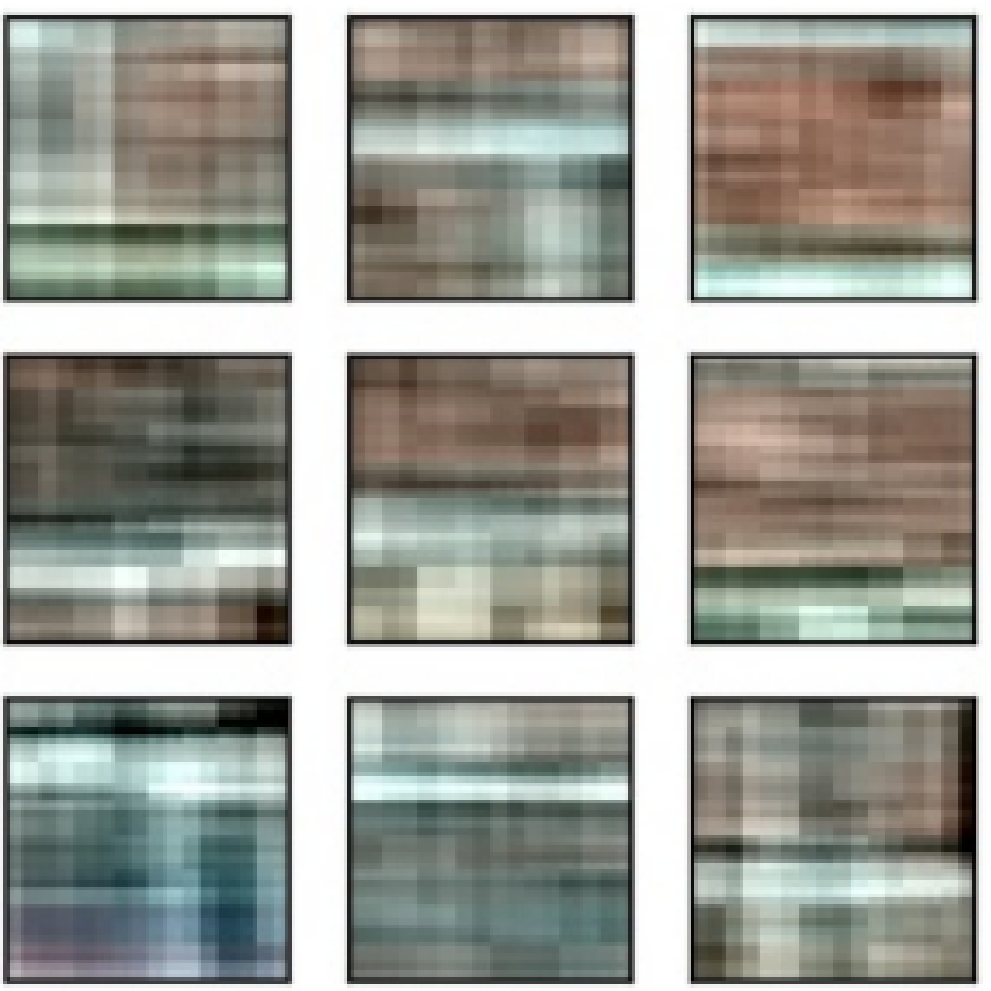} }}%
    \caption{Generated \emph{brick\_wall}  patches using different threshold $\gamma$ values in SDR.}%
    \label{fig:sdr}%
\end{figure*}

\begin{table}[htb]
\caption{The number of parameters of TGHop, under the setting of $\gamma=0.01$, 
$N=50$ $K_1=9$, $K_2=22$, $D_r=909$, $F=2,518$ and $W=200$.}\label{table:modelsize}
\begin{center}
\begin{tabular}{lll} \\ \hline
Module  &  Equation & Num. of Param.  \\ \hline
Transform - stage 1 & $12K_1+1$ &  109 \\
Transform - stage 2 & $4K_1K_2+K_1$ &  801 \\
Core - SDR & $64D_r$ & 58,176 \\
Core - ICHM(i) & $N$ &  50  \\
Core - ICHM(ii) & $FD_r$ &  2,288,862  \\  
Core - ICHM(iii)  & $3F+256W$ & 58,754 \\  \hline
\bf{Total}  & & \bf{2,406,752}  \\ \hline
\end{tabular}
\end{center}
\end{table}

The above equations are summarized and an example is given in
Table~\ref{table:modelsize} under the experiment setting of $N=50$
$K_1=9$, $K_2=22$, $D_r=909$, $F=2,518$ and $W=200$. The model size of
TGHop is 2.4M. For comparison, the model sizes
of~\cite{gatys2015texture} and \cite{ustyuzhaninov2017does} are 0.852M
and 2.055M, respectively.  A great majority of TGHop model parameters
comes from ICHM(ii). Further model size reduction without sacrificing
generated texture quality is an interesting extension of this work.

\begin{table}[htb]
\begin{center}
\caption{The reduced dimension, $D_r$, and the model size as
a function of threshold $\gamma$ used in SDR.} \label{table:gamma}
\begin{tabular}{ccc} \\ \hline
$\gamma$ & $D_r$ & Number of Parameters\\ \hline
0 & 1408 & 3.72M \\ 
0.0005 & 1226 &  3.26M\\ 
0.005 & 1030 &  2.74M\\ 
0.01 & 909 &  2.41M\\ 
0.02 & 718 &  1.88M\\ 
0.03 & 553 &  1.43M\\ 
0.04 & 399 &  1.00M\\ 
0.05 & 289 &  0.69M\\ 
0.1 & 102 &  0.19M\\ \hline
\end{tabular}
\end{center}
\end{table}

As compared with~\cite{lei2020nites}, SDR is a new module introduced in
this work. It helps remove correlations of spatial responses to reduce
the model size.  We examined the impact of using different threshold
$\gamma$ in SDR on texture generation quality and model size with
\emph{brick\_wall} texture. The same threshold is adopted for all
channels to select PCA components.  The dimension of reduced space,
$D_r$, and the cluster number, $N$, are both controlled by threshold
$\gamma$, used in SDR.  $\gamma=0$ represents all 64 PCA components are
kept in SDR.  We can vary the value of $\gamma$ to get a different
cluster number and the associated model size.  The larger the value of
$\gamma$, the smaller $D_r$ and $N$ and, thus, the smaller the model
size as shown in Table \ref{table:gamma}. The computation given in
Table \ref{table:modelsize} is under the setting of $\gamma=0.01$. 

A proper cluster number is important since too many clusters lead to
larger model sizes while too few clusters result in bad generation
quality.  To give an example, consider the \emph{brick\_wall} texture
image of size $256\times256$, where the dimension of $S_2$ is $8\times 8
\times 22=1408$ with $K_2=22$.  We extract 12,769 patches of size
$32\times32$ (with stride 2) from this image. We conduct experiments
with $N=$50, 80, 110 and 200 clusters and show generated patches in
Fig.~\ref{fig:numclu}. As shown in (a), 50 clusters were too few and we
see the artifact of over-saturation in generated patches. By increasing
$N$ from 50 to 80, the artifact still exists but is less apparent in
(b). The quality improves furthermore when $N=100$ as shown in (c). We
see little quality improvement when $N$ goes from 100 to 200.
Furthermore, patches generated using different thresholds $\gamma$ are
shown Fig.~\ref{fig:sdr}. We see little quality degradation from (a) to
(f) while the dimension is reduced from 1408 to 553. Image blur shows up
from (g) to (i), indicating that some details were discarded along with
the corresponding PCA components. 

\section{Conclusion and Future Work}\label{sec:conclusion}

An explainable, efficient and lightweight texture generation method,
called TGHop, was proposed in this work. Texture can be effectively
analyzed using the multi-stage c/w Saab transforms and expressed in form
of joint spatial-spectral representations. The distribution of sample
texture patches was carefully studied so that we can generate samples in
the core.  Based on generated core samples, we can go through the
reverse path to increase its spatial dimension.  Finally, patches can be
stitched to form texture images of a larger size. It was demonstrated by
experimental results that TGHop can generate texture images of superior
quality with a small model size and at a fast speed. 

Future research can be extended in several directions. Controlling the
growth of dimensions of intermediate subspaces in the generation process
appears to be important. Is it beneficial to introduce more intermediate
subspaces between the source and the core? Can we apply the same model
for the generation of other images such as human faces, digits,
scenes and objects? Is it possible to generalize the framework to image
inpainting? How does our generation model compare to GANs? These are all
open and interesting questions for further investigation.

\section*{Acknowledgment}

This research was supported by a gift grant from Mediatek. Computation
for the work was supported by the University of Southern California's
Center for High Performance Computing (hpc.usc.edu). 

\bibliographystyle{unsrt}
\bibliography{references.bib}

\end{document}